\documentclass{article}
\usepackage[utf8]{inputenc}

\usepackage{amsmath}
\usepackage{amsthm}
\usepackage{amssymb}

\usepackage{xcolor} 

\usepackage{microtype}
\usepackage{graphicx}
\usepackage{subfigure}
\usepackage{booktabs}
\usepackage{tabularx}

\usepackage{hyperref}

\newtheorem{property}{Property}
\newtheorem{definition}{Definition}
\newtheorem{problem}{Problem}

\usepackage[accepted]{icml2021} 

\icmltitlerunning{Towards Rigorous Interpretations: a Formalisation of Feature Attribution}

\begin{document}

\twocolumn[
\icmltitle{Towards Rigorous Interpretations: a Formalisation of Feature Attribution}

\begin{icmlauthorlist}
\icmlauthor{Darius Afchar}{deezer,lip6}
\icmlauthor{Romain Hennequin}{deezer}
\icmlauthor{Vincent Guigue}{lip6}
\end{icmlauthorlist}

\icmlaffiliation{deezer}{Deezer Research, Paris, France}
\icmlaffiliation{lip6}{LIP6, Paris, France}

\icmlcorrespondingauthor{Darius Afchar}{research@deezer.com}

\icmlkeywords{Machine Learning, ICML}

\vskip 0.3in
]
\printAffiliationsAndNotice  

\begin{abstract}
Feature attribution is often loosely presented as the process of selecting a subset of relevant features as a rationale of a prediction. Task-dependent by nature, precise definitions of "relevance" encountered in the literature are however not always consistent. This lack of clarity stems from the fact that we usually do not have access to any notion of ground-truth attribution and from a more general debate on what good interpretations are. In this paper we propose to formalise feature selection/attribution based on the concept of relaxed functional dependence. In particular, we extend our notions to the instance-wise setting and derive necessary properties for candidate selection solutions, while leaving room for task-dependence. By computing ground-truth attributions on synthetic datasets, we evaluate many state-of-the-art attribution methods and show that, even when optimised, some fail to verify the proposed  properties and provide wrong solutions.
\end{abstract}


\section{Introduction}

As the adoption of intelligent algorithms of growing complexity is becoming ubiquitous in our everyday lives,
concerns have consequently emerged about the lack of transparency and need for interpretability of these methods \cite{gdpr}. Interpretability is unfortunately somewhat ill-defined and ill-evaluated  \cite{doshi2017towards, lipton2018mythos}, partly because a wide range of concepts are encompassed under the same label. One can think of the protean purposes of interpretations: \textit{informativeness}, \textit{causality}, \textit{fairness}, \textit{interactivity}, \textit{trust}, etc \cite{tintarev2007survey, arrieta2020explainable}.  Nonetheless, there is a consensus on the fact that interpretations stem from a notion of \textit{incompleteness} and aim at boosting \textit{human understandability}. But viewing understandability in an holistic manner requires controlling and disentangling every aspect of a model prediction, ranging from how the data is inherently structured, to what priors are induced by a model architecture, to what impact can a design choice to present explanations have on a target audience and in a particular setting. 

\begin{figure}[h]
    \centering
    \includegraphics[width=\linewidth]{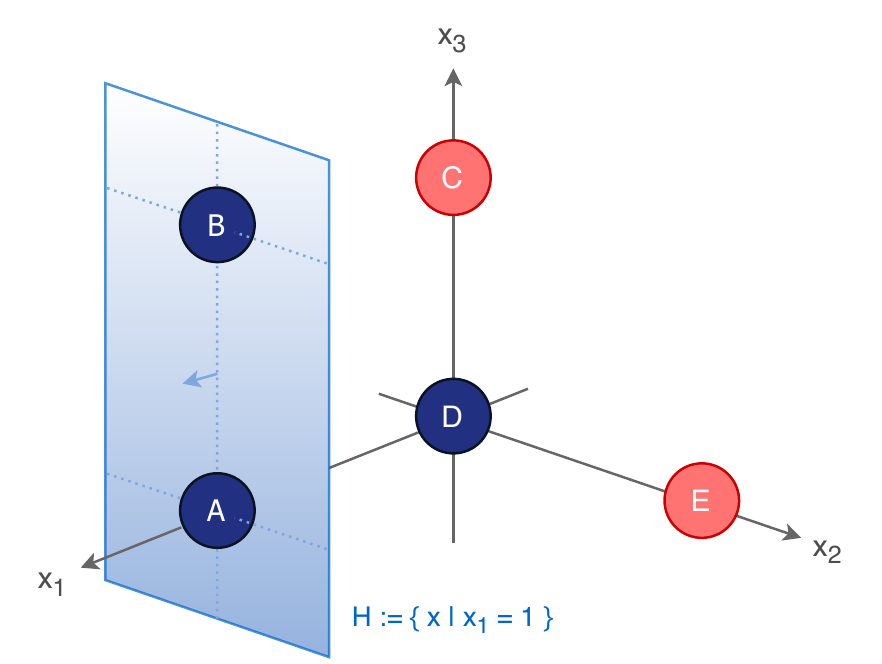}
    \caption{\textbf{Intuition of the formalisation} Example prediction task with five points $p_A, ..., p_E$ in $\mathbb{R}^3$ and binary labels blue/red.
    Note that, to correctly label point $p_A$ as \textit{blue}, it is sufficient to know that the point has its coordinate $(p_A)_1 = 1$.
    Indeed, the incomplete view that $x_1 = 1$, may lead to confuse points $p_A$ and $p_B$, but not the determination of their label (\textit{blue}). We say that $p_A$ functionally depends on $X_1$. Since all points in $H$ symmetrically have the same label, they also share the same dependence in $X_1$; by comparison, this does not hold for other points at $x_1 = 0$.
    We will see that this symmetry argument is a necessary property for any instance-wise feature selection candidate solution.
    }
    \label{fig:front}
\end{figure}

This may explain why many methods have resorted to proxy measures of interpretability and have proposed list of general requirements for interpretations - \textit{e.g.} \cite{ribeiro2016should, NIPS2017_7062, sundararajan2017axiomatic}, that are sometimes confirmed by user-studies: \textit{e.g.} sparsity is widely considered a general desiderata of interpretation. This process is not always successful \cite{rudin2019stop}. In fact, many recent works \cite{adebayo2018sanity, serrano2019attention, kindermans2017reliability, dombrowski2019explanations, sixt2020explanations, kumar2020problems}
tend to suggest that well-established interpretation methods may not provide much understandability after further inspection, while being coherent with their self-defined interpretation criteria.
Additionally, \citet{kaur2020interpreting} showed that several popular methods may be misused by practitioners with lack of consideration for methods' assumptions relevance or requirements or application domain, and thus be prone to confirmation biases.
Such blunders are not new in the field of interpretable machine learning, which is why \citet{doshi2017towards} had advocated for rigorous formalisation so as to avoid any subjective definition, vague evaluations and practitioners misuses, such as what had already been done in the subfields of fairness or privacy.

In this paper we propose to formalise a popular class of interpretation methods that we find lacks clarity: \textbf{feature attribution}.
Feature attribution/importance aims at providing a rationale for the association of target values to input instances; where target values may correspond to a model's predictions -- enabling the inspection of its behaviour, or observed true labels -- to interpret data.
To do that, all attributions tasks can be decomposed into two subproblems: \textbf{(1)} providing a scoring function that represents the \textit{responsibility} of a feature or group of features in the association to a given value, then \textbf{(2)} returning a parsimonious subset of features as a rationale of the association, using the scores. The rationale can either apply to all instances -- \textit{global attribution}, allowing to discard noisy and redundant features \cite{tibshirani1996regression}, or be computed locally -- \textit{instance-wise attribution}.
The concept of \textit{responsibility} is however task-dependent and varies widely between methods, and the relevance of the returned minimal features is sometimes ill-evaluated, if evaluated at all.
In particular in the instance-wise setting, and unlike global attribution, we will show that checking prediction performances from selected features is not sufficient to ensure that the correct rationale was found.

That said, ground-truth knowledge of input responsibility is not usually available in any form for collected data.
Furthermore, evaluations on real data often come with the hardship of disentangling interpretation errors from prediction errors \cite{dinu2020challenging}.

That is why we propose to study in detail an informed scenario, for which we know everything about the input distribution $p_X$ and target distribution $p_{Y\mid X}$. Specifically, we generate synthetic supervised tasks and abstract models from the task by replacing them with optimal distributions or mappings\footnote{For instance, $\mathbb{E}_{Y | X}[Y | X]$ for regression tasks with normal priors and $\arg\max_c p_{Y \mid X}(y = c \mid X)$ for categorical tasks.}. Doing so, we are able to derive ground-truth rationales and critically assess the interpretation capabilities of many attribution methods. Our vision is that if a method fails at providing relevant attributions given this ideal and noise-controlled distribution of the data, this should be worrisome for real-world applications.

Our contributions are the following:
\begin{enumerate}
    \item We propose a formalisation of selection and attribution based on functional dependence and derive necessary properties to extend them to the instance-wise setting;
    \item We rigorously evaluate feature selections of many state-of-the-art methods on generated data and show that only a few of them achieve satisfying performances; 
    \item We show that our proposed necessary properties allow to evaluate estimated selections quality without having access to ground-truth solutions.
\end{enumerate}

\begin{figure*}[h]
    \centering
    \includegraphics[width=\textwidth]{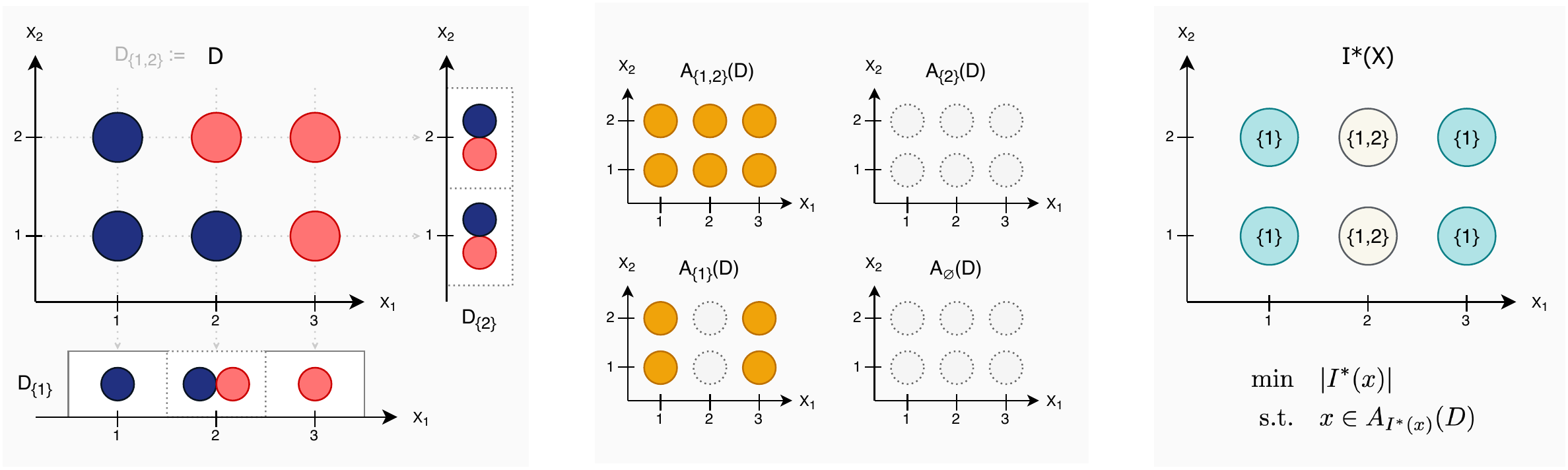}
    \caption{\textbf{Example of instance-wise selection derivation}
    From left to right: we are given a relation $D$ from $\mathcal{X} = [3] \times [2]$ to $\mathcal{Y} = \{ \textrm{blue}, \textrm{red} \}$, for simplicity, we assume that it defines unique associations (\textit{ie.} $D$ is a function), we compute its associated projected relations $D_I$; then its functionality domains $A_I(D)$; and finally derive the instance-wise selection solution $I^*(x)$. 
    }
    \label{fig:intro}
\end{figure*}


\section{Feature attribution formalisation}

We start by defining some notations. As mentioned, we study a supervised prediction interpretation setting: let us denote by $x \in \mathcal{X}$ an input sample and $y \in \mathcal{Y}$ its associated label or continuous value. We suppose $\mathcal{X} \subseteq \mathbb{R}^n$ and denote $[n] = \{1, ... n\}$ the set of input indexes. The attribution problem is that for a given sample $x$ and for all subsets $I \subset [n]$, we first want to estimate a value $\textrm{attr}_I(x)$ that represents the \textit{responsibility} of $(x_k)_{k \in I}$ in the observed association of $x$ to $y$, and then return a minimal responsible subset using all the values $(\textrm{attr}_J(x))_{J \subset [n]}$.

The issue is that responsibility, sometimes referred to as \textit{relevance} or \textit{importance}, is ill-defined. There are however two principles that are shared across all attribution methods that will guide us in our formalisation. First, since interpretations depend on their application field and target audience, \textbf{responsibility is task-specific (\textit{P1})}. For instance, it is sometimes relevant to have a notion of negative responsibility -- \textit{e.g.} in sentiment prediction tasks to find words that flip the meaning of a sentence; and sometimes not -- \textit{e.g.} for a recommender system using an implicit feedback dataset where negative interactions are not meaningful \cite{hu2008collaborative}.
The second principle lies in the binary distinction between \textit{null} and \textit{non-null} responsibilities: a null value indicates a subset of variables that has nothing to do with the association of $x$ to $y$; a non-null one does, to some task-specific extent. \textbf{Responsibilities should enable to distinguish contributing and non-contributing features (\textit{P2})}. Splitting input features into a minimal subset of contributing features versus non-contributing others is called the \textbf{feature selection problem} \cite{natarajan1995sparse, blum1997selection}. We argue that selection should always be implied by attribution, and by contraposition, that an attribution method that does not allow to return a correct selection solution should be questioned.

In the rest of the section, we first formalise the notion of \textit{contributing subset of features} from (\textit{P2}) using the concept of functional dependence. In particular, we will extend our notion to the instance-wise setting which lacks formalism. Then, for (\textit{P1}), we propose to see \textit{responsibility} as its probabilistic relaxation and derive task-specific examples.

Now, to properly define what contribution means, we come back to the definition of a function and answer the following question: \textit{what does it mean for a function to depend or not on a set of variables?}

\subsection{Background on functionality}

In set theory, the notion of function is built from the concept of \textit{binary relation}. We adapt the definition and notation from \citet{hamilton1982numbers}.

\begin{definition}[Binary relation]
A binary relation from a set $\mathcal{X}$ to a set $\mathcal{Y}$ is a subset of the Cartesian product of the two sets.
If $R$ is such a relation and $(x, y) \in R$, we say that $x$ is related to $y$ and for convenience we may write $xRy$. 
\end{definition}

What differentiates a relation from a function is that multiple outcomes can be in the image of a single input element of a relation. The second difference is that some points from $\mathcal{X}$ may not have been related to any point in  $\mathcal{Y}$. Hence the following definition:

\begin{definition}[Function]
\label{def:function}
A partial function $f$ is a binary relation that is single-valued. For all $x \in \mathcal{X}$, $y, z \in \mathcal{Y}^2$:
\begin{equation*}
    ((x,y) \in f) \; \wedge \; ((x,z) \in f) \; \Rightarrow \; y = z    
\end{equation*}
To obtain a function, we additionally require this partial function $f$ to be left-total:
\begin{equation*}
    \forall x \in \mathcal{X}, \; \exists y \in \mathcal{Y}, \; (x, y) \in f
\end{equation*}
When these two conditions are met, we can write the familiar expression $f(x)$ that denotes for all points of $\mathcal{X}$ the existing and unique element $y \in \mathcal{Y}$ such that $xfy$.
\end{definition}

These two definitions are the starting point of our formalisation. We consider a given dataset of samples and associated labels. We want to express it as a dependence between the given input and associated labels.
By definition, a dataset induces a binary relation between an input set $\mathcal{X}$ and a target set $\mathcal{Y}$ (continuous or discrete, it does not matter at this point). We denote it $D$. Without loss of generality, we assume $D$ to be left-total, or reduce $\mathcal{X}$ accordingly. The single-value condition in definition \ref{def:function} tells us when it is possible or not to uniquely assign a target label/value to a point in space, hence creating a functional dependence, \textit{i.e.} given a dataset, this point is always related to a specific target, it \textit{implies} it. For a given binary relation $R$, we define the subset of the domain $\mathcal{X}$ such that this condition is met:
\begin{equation*}
    A(R) = \{ x \in \mathcal{X} \mid \forall y,z \in \mathcal{Y}^2, \; xRy \wedge xRz \Rightarrow y = z \} 
\end{equation*}
By construction, our dataset $D$ with its domain restricted to $A(D)$ is a function, meaning that points of $D$ are \textit{uniquely associated} on $A(D)$. By contrast, all points in $\bar{A}(D)$ are such that multiple target labels are related to a single input, the information given by the sole point position is intrinsically not sufficient to predict or assign a label. This is aside from the probabilistic considerations we will have in \ref{sec:relaxation}.

\subsection{Subset functionality and selection}

For selection though, we are interested in \textit{defining dependence to only a subset among the input dimensions}. To do that, we first ignore some features by setting them to zero. There, it is convenient to use canonical projections. We denote $\mathcal{X} \subseteq \mathbb{R}^n$, $(\vec{e_1},... \vec{e_n})$ the canonical base of $\mathbb{R}^n$ and for a subset of indices $I \subset [n]$ the canonical projection on those indices $P_I(x) = \textrm{proj}(x, \{ \vec{e_k} \mid k \in I \})$, and then define for a relation $R$, its relation with projected domain $R_I$:
\begin{equation*}
    R_I = (P_I \times \textrm{Id}_{\mathcal{Y}})(R)
\end{equation*}
For our dataset $D$, $D_I$ is the dataset such that all input features with indices not in $I$ are set to zero\footnote{Because we know the subset $I$ we project on, we can distinguish a \textit{data} zero in $I$ from the \textit{ignoring} zeros of $\bar{I}$.}. As a result, multiple points in the domain of $D$, with potentially different labels, may be collapsed into a single representative in $D_I$, thus killing the functionality property they may have verified in $A(D)$. On the point-wise level, the construction of a projected relation $R_I$ implies that if $xRy$ then $(P_Ix)R_Iy$, and reciprocally if $x_IR_Iy$, there exists an antecedent $x$ such that $xRy$ and $P_Ix = x_I$.
We refer to figure \ref{fig:intro} for a simple example to reason about the different concepts introduced in this section.

We now extend the previous definition of functional domain $A$ to the case where we only consider subsets of features.
\begin{definition}
\label{def:local_dependance}
    For a given relation $R \subset \mathcal{X} \times \mathcal{Y}$, a subset of indices $I \subset [n]$ and $R_I$ its projection to $I$, $A_I(R) \subset \mathcal{X}$ is the subset such that for all $x \in \mathcal{X}$, $y, y' \in \mathcal{Y}$, $x_I = P_Ix$,
    \begin{equation*}
    x \in A_I(R) \Leftrightarrow (x_IR_Iy \wedge x_IR_Iy' \Rightarrow y = y')
    \end{equation*}
    Or equivalently, $x$ is in $A_I(R)$ if and only if
    \begin{equation*}
        \forall x' \in \mathcal{X} \; \textit{s.t.}\, P_Ix' = P_Ix, \; xRy \wedge x'Ry' \Rightarrow y = y'
    \end{equation*}
\end{definition}
\begin{proof}
$(P_Ix' = P_Ix = x_I) \wedge (xRy) \wedge (x'Ry') \Leftrightarrow (x_IR_Iy) \wedge (x_IR_Iy') $
\end{proof}

By construction, $D_I$ with its domain restricted to $P_I(A_I(D))$ is a function. Or said differently, for a given subset of indices $I$, for all points $x \in A_I(D)$, a target label $y$ can be uniquely associated to $x$ by the mere knowledge of its subset $I$ of features. By comparison to definition \ref{def:function}, the only added condition is that the single-valueness must be verified not only by $x$ but also all points with the same projection as $x$ on $I$.


Now, once we have computed all $2^n$ domain subsets $A_I(D)$, the selection problem is formulated as the task of finding minimal subsets of input indices that all points functionally depend on. Which leads to two possible settings:
\begin{problem}[Global subset selection]
Given a relation $R$, find a subset of indices $I^* \subset [n]$ that minimises
\begin{align*}
        \min\limits_{J \subset [n]} \; & \textrm{Card}(J) \\
        \textrm{s.t.} \quad & \forall x, \; x \in A_J(R)
\end{align*}
\end{problem}

\begin{problem}[Instance-wise subset selection]
Given a relation $R$, for all $x \in \mathcal{X}$, find a local subset of indices $I^*(x) \subset [n]$ that minimises
\begin{align*}
        \min\limits_{J \subset [n]} \; & \textrm{Card}(J) \\
        \textrm{s.t.} \quad & x \in A_J(R)
\end{align*}
\end{problem}
Note that it is not assured that these minima are unique, which is not problematic and rather natural, for instance when some input features are correlated.

Our derived definition of dependence/contribution and \textit{global} selection coincides with \citet{blum1997selection}. In the rest on the paper, we study its \textit{instance-wise} extension, for it is the most difficult case with the largest risk of providing degenerate explanations if not done carefully.


\subsection{Necessary properties of instance-wise dependence}
\label{sec:necessary}
The above definitions allow us to derive properties a given instance-wise selection solution $\hat{I}(x)$ should verify.

\begin{property}[Complementary dependence]
\label{prop:hyperspace}
If a point depends on a subset of indices, all point in directions in the complement of this subset have the same dependence : for $x \in \mathcal{X}$, if there exists $I \subset [n]$ such that $x \in A_I(R)$, then for all $ x' \in \mathcal{X}$ such that $P_Ix' = P_Ix$, one has $x' \in A_I(R)$.
\end{property}
\begin{proof}
$[(P_I(x'),y') \in R_I] \wedge [(P_I(x'),y'') \in R_I] = [(P_I(x),y') \in R_I] \wedge [(P_I(x),y'') \in R_I] \Rightarrow y' = y''$
\end{proof}

This property is illustrated in figure \ref{fig:front}.
We will see in the experiment section \ref{sec:exp} that this property is not verified by some widely used attribution methods.

\begin{property}[Dependence hierarchy]
\label{prop:parents}
Any point that depends on a subset also depends on its parent subsets : $I \subset J \Rightarrow A_I(R) \subset A_J(R)$.


\end{property}
\begin{proof}
$R_I = ((P_I \times \textrm{Id}_{\mathcal{Y}}) \circ (P_J \times \textrm{Id}_{\mathcal{Y}}))(R)$, thus $(P_JxR_Jy) \wedge (P_Jx'R_Jy') \Rightarrow (P_IxR_Iy) \wedge (P_Ix'R_Iy') \Rightarrow y = y'$
\end{proof}


\subsection{Attribution as relaxed functional dependence}
\label{sec:relaxation}

We have formalised the notion of binary feature contributions for the selection task in quite an unrealistic case where we could find a perfect dependence. We now propose to \textit{frame attribution values as its probabilistic relaxation}.
Indeed, there are several reasons we may want to adopt a probabilistic framework and relax functional dependence:
\begin{itemize}
    \item Real-data is noisy, we only have access to a sample of it, and may wish to control a certainty of dependence;
    \item For continuous $\mathcal{Y}$, we may tolerate having several outcomes for $x \in \bar{A}_I(R)$ but that are close to one another; and for categorical $\mathcal{Y}$, a small stochasticity of label;
    \item Generally, we want to accurately model probable associations of input and target label/values while minimising the weight of rare and out-of-distribution points.  
\end{itemize}

Instead of a dataset $D$, we now consider probabilistic densities $p_X$ and $p_{Y\mid X}$ on $\mathcal{X}$ and $\mathcal{Y}$ with their usual associated input and target random variables $X$ and $Y$. We relax our notion to \textit{approximate functional dependence}. 
At that point, we have to consider task-dependency as there is no one-relaxation-fits-all rule (\textit{P1}). Attribution values should however still allow to differentiate between relevant and non-relevant subset of features to be meaningful (\textit{P2}). As a general framework, we first define an attribution relaxation $\textrm{attr}_I(x)$ for all subsets $I \subset [n]$ and all samples $x \sim X$, and we then create the link to selection with a comparison to a chosen threshold parameter $\eta$. For instance, we could choose that all subsets of features with absolute attribution value higher than $\eta$ should be selected. We can not define an encompassing comparison mechanism, the implication mechanism from attribution to selection is part of the relaxation elaboration and directly translates the meaning of the degree of approximation we choose with $\eta$. We give some examples of attribution relaxation to clarify this framework.

\textbf{Regression setting}
Let $Y$ be continuous, \textit{e.g.} $\mathcal{Y} = \mathbb{R}$, and the function we want to interpret be the mean mapping $f(x) = \mathbb{E}[Y \mid X = x]$.
To define an instance-wise responsibility measure $g_I$ that will imply functional dependence on $I$, we can use the conditional variance:
\begin{align}
    \label{measure:variance}
    g_I(x) & = \textrm{Var}_{X \mid X_I}[ Y \mid X_I = P_I(x) = x_I ]
\end{align}
where $X_I$ denotes the projected random variable $P_I(X)$. We verify that $g_I(x) = 0$ if and only if for all samples $(x', y')$ such that $P_Ix' = x_I$, the associated value $y'$ is equal to the conditional mean $\mathbb{E}_{X_{\bar{I}} \mid X_I}[Y \mid X_I = x_I]$, hence verifying $x \in A_I(f)$ and thus (\textit{P2}) in the perfect setting.

In the literature, it is more usual that attribution values near zero denote independence to a subset. To do that, we could use the reciprocal notion of \textit{precision}: $\textrm{attr}_I(x) = 1 / g_I(x)$. When the precision is low, the samples with common features on the indices $I$ are spread, it is thus not possible to assign a value that will be representative enough of these points. When precision is high, the mean value will be a relevant predictor of the points, we can state that we have a dependence to $I$ \textit{with a given precision/variance}.

With this measure and for a given variance threshold $\eta$, we have obtained \textbf{approximated functionality domains}:
\begin{equation}
    A_I^\eta(f) = \{ x \in \mathcal{X} \mid  |\textrm{attr}_I(x)| \geq 1/\eta \}
\end{equation}
again, we verify that $A_I^0(f) = A_I(f)$ (\textit{P2}).

To fix ideas through a simple application example, let us consider a bidimensional uniform input $X = (X_1, X_2)$ on $\mathcal{X} = [-1, 1]^2$, and $Y$ such that,
\begin{align*}
    & p_X = p_{X_1}p_{X_2} = 1/4 \\
    & Y = X_1 + \alpha X_2, \quad | \alpha | < 1 
\end{align*}
which corresponds to a deterministic identity mapping from $X_1$ to $Y$ with a small tilt effect from $X_2$ with coefficient $\alpha$. Then, for all $x_1 \in [-1, 1]$,
\begin{equation*}
    \textrm{Var}_{X_2 \mid X_1}[Y \mid X_1 = x_1] = \frac{1}{2} \int\limits_{-1}^{1} (\alpha t)^2 {dt} = \alpha^2 / 3
\end{equation*}
for a given variance threshold $\eta$, the attribution measure \eqref{measure:variance} states that if $\alpha \leq \sqrt{3\eta}$, the target variable $Y$ can be approximated with $Y' = X_1$, \textrm{i.e.} $X_2$ is ignored and only $X_1$ is responsible for $Y$.

Similarly, let us have $Y = X_1 + \epsilon$, with $\epsilon$ a noise variable following $\mathcal{N}(0, \sigma^2)$. Because of the noise, there is no region of the domain where samples of $Y$ can be uniquely determined on a set of variables. But when $\sigma^2 \leq \eta$, the noise can be ignored and this distribution can be approximated by the univariate distribution of $Y' = X_1$ with variance $\eta$.

With the attribution measure \eqref{measure:variance}, we were able to relax dependence to a probabilistic framework allowing to control noise and small feature effects, and yield approximate feature contribution. Our choice of relaxation through the conditional variance works well when $Y$ is assumed to follow a normal law $\mathcal{N}(\mu(X), \sigma(X)^2)$. This is of course not the only possible attribution measure, in particular if we want to study more than the mean effects $f$ we chose.




\textbf{Classification setting}
When $Y$ takes values in the set of $n$ labels $c_1, \dots c_n$. It seems natural to define an attribution measure as the probability of assigning the label with maximum probability.
\begin{align}
    P^c_I(x) & = \mathbb{P}( Y = c \mid X_I = P_Ix ) \nonumber \\
    \textrm{attr}_I(x) & = \max\limits_c P^c_I(x) \label{measure:prob} \\
    A_I^\eta(f) & = \{ x \in \mathcal{X} \mid  \textrm{attr}_I(x) \geq 1 - \eta \}
\end{align}
The attribution value is bounded between $1/n$ (uniform) and $1$ (deterministic label). These responsibilities have a nice interpretation since they directly represents the proportion of samples in the same class when conditioning on the variables in $I$. Adjusting $\eta$ also means that we control the error on the prediction of a class for these samples.

In the perfect setting, for $\eta = 0$ we check that $x \in A_I^0(f) \Rightarrow \textrm{attr}_I(x) = 1 \Rightarrow x \in A_I(f)$, thus (\textit{P2}). In the imperfect setting, our function $f$ under study is noisy, \textbf{the goal is to tune $\eta$ to maximise the verification of (\textit{P2})}, which we will evaluate in section \ref{sec:exp}.

Alternatively, it may be more relevant to take in consideration all labels probabilities with an entropy measure:
\begin{align}
    \textrm{attr}_I(x) & = 1 - \sum_c \frac{P^c_I \ln(P^c_I)}{\ln(1/n)} \label{measure:entropy}
\end{align}
We have normalised the entropy to obtain a value between 0 (uniform label distribution) and 1 (deterministic label).


\section{Related methods}

We present classic and state-of-the-art selection/attribution methods in the light of the formalism we propose, and with a specific focus on instance-wise methods. Due to size constraints, it is impossible to present all variations of assumptions and clever solutions of these methods, we will thus only present four general ideas that, we think, constitute the bulk of research on instance-wise feature attribution.

\subsection{Mixture of restricted experts}

The first thing we have to mention is that the attribution relaxation \eqref{measure:variance} we introduced in the context of regression is strongly inspired by the success of the classical \textit{analysis of variance} diagnostics and its more recent formulation of \textit{weighted functional ANOVA} \cite{hooker2007generalized} that decomposes $\mathcal{L}_2$ functions into the sum of all $n$-variate subfunctions under a hierarchical orthogonality constraint, weighted by the data distribution. 
Given that one takeaway of our paper will be that we have to consider the full input distribution for relevant interpretations, not just local information, we should have been happy with weighted fANOVA.
Specifically, one key consequence of fANOVA is that the overall variance can be decomposed as a sum of variance from each subfunction, and hence each input subset. However, this decomposition is made identifiable through an \textit{integration-to-zero} constraint on the subfunctions, allowing to formulate global selection criteria but not to distinguish the non-null instance-wise contributions we seek from any centering effects (see Supplementary A). 

Another idea, similar in spirit to fANOVA, is to try to directly learn a mixture of $n$-variate functions. Since there is a potential exponential number of subfunctions, one approximation making training tractable is to consider only summed univariate contributions -- \textit{e.g.} \textit{GAM} \cite{hastie1990generalized}; or interactions up to a fixed order -- \textit{e.g.} \textit{GA\textsuperscript{2}M} \cite{lou2013accurate}, \textit{NIT} \cite{tsang2018neural}; or with a fixed structure -- \textit{e.g.} \textit{Archipelago} \cite{tsang2020does}, \textit{InterpretableNN} \cite{afchar2020making}.
The key advantage of mixture models is that they disentangle the different orders of interaction effects. In our formulation of dependence, no distinction can for instance be made between $f(x) = x_1 + x_2$ and $f(x) = x_1x_2$ with a uniform input distribution. This may be useful in some applications. But conversely, and beyond the trivial limitation that these models provide solutions within a restricted candidate space, additive models strongly suffer from an identifiability issue and can produce contradictory interpretations. Identifiability can be achieved with fANOVA-like regularisation \cite{lengerich2019purifying}, but we have argued that this does not allow to obtain exact attribution in an instance-wise setting. This effect gets worst with high-order interactions and redundant or correlated features. Meanwhile, our attribution formalisation allows to distinguish multiple possible candidate solutions, hence isolating redundancies, but at the cost of interaction hierarchical decomposability. We may assert that both approaches are complementary.

\subsection{Proxy models}
\label{sec:proxy_methods}
A large body of work on instance-wise attribution circumvents the above tractability issue by providing proxy measurements of attribution. Two large class of methods are \textbf{gradient-based} analysis -- \textit{e.g.} saliency methods \cite{simonyan2013deep}, \textit{SmoothGrad} \cite{smilkov2017smoothgrad}, ... ; and \textbf{baseline-comparison} methods -- \textit{e.g.} \textit{LIME} \cite{ribeiro2016should}, \textit{SHAP} \cite{NIPS2017_7062}, ... ; the line between these two classes is fuzzy -- \textit{e.g.} \textit{Integrated Gradient} \cite{sundararajan2017axiomatic}, \textit{Expected Gradient} \cite{erion2019learning}. Again, we will not discuss the profusion of variations but only their general spirit. For good meta-analysis on a unification of these methods, we recommend \cite{covert2020feature} and \cite{sundararajan2020many}. Nevertheless, the underlying principle behind the computation of a gradient as an indication of feature contribution can be found in its simplest form in \citet{friedman2008}. In substance, it says that \textit{a function $F(x)$ is said to exhibit an interaction between $k$ variables with indexes $I = (i_1, \dots i_k)$ if $\mathbb{E}_X[{\partial^k F}/{\partial x_{i_1} \dots \partial x_{i_k}}]^2 > 0$}, meaning that the difference in value of $F(x)$ as a result of changing some variables of $I$ depends on the remaining variables of $I$. Beyond noise considerations that may create nuisance interactions, this approach is rather sound for global selection. Problems occur in its extension to the instance-wise setting when $\mathbb{E}_X$ is dropped without any further considerations. This is the foundation of saliency methods and subsequent papers have focused on providing gradient estimates that proved robust to noise. To adopt the same formalism as before, we could write those gradient-based selection measures in the general form:
\begin{equation}
    \label{measure:gradient}
    G_I(f) = \{ x \in \mathcal{X} \mid ({\partial^{|I|} f(x)} / {\partial X_I})^2 > 0 \}
\end{equation}
with $f$ a function. For a relaxed formulations for attribution, many aspects have to be considered to provide a relevant estimate for the derivatives for a given task, we will not discuss them here and assume an ideal favorable setting where this measure is available. 

Baseline-comparisons methods, in the spirit of counterfactual reasoning, determine the extent to which a function output differs from an output considered "neutral" -- the baseline. Many choices exist to model the baseline, a common one is to estimate a conditional expectation. We may formalise them in the general form:
\begin{equation}
    \label{measure:baseline}
    C_I(f) = \{ x \in \mathcal{X} \mid f(x) \neq \mathbb{E}[f(X) \mid X_I = P_Ix ] \}
\end{equation}
choosing another baseline, as $f(X_{I}, \mathbb{E}_{\bar{I}}(X_{\bar{I}} ) ) $ \cite{NIPS2017_7062} does not change our discussion.

To link these two subsets with previous notions, we introduce the following subset of $\mathcal{X}$:
\begin{equation}
    \label{eq:B}
    B_I(f) = \{ x \mid \exists x', \; P_{\bar{I}}x' = P_{\bar{I}}x, \; f(x) \neq f(x')  \}
\end{equation}
\textit{i.e.} the set of points $x$ for which when fixing the $I$ features, there is still a alternate value for $f$. This notion is reminiscent of the functionality property in the subsets $(A_I)$, and indeed we have the trivial connection $B_I = \overline{A_{\bar{I}}}$. Then, for gradient-based methods, having a finite non-null gradient implies that there exists a neighborhood such that there exists distinct values for $f$, and hence $G_I \subset B_I$. But gradient methods miss some cases, for instance if $f$ is constant in the neighborhood of $x$ but vary further away, $x$ will not be included in $G_I$. Similarly, we have $C_I \subset B_I$: to find a probable point that is different from an average, there must exists points with different value that counterweight its deviation from the mean. Note that the case of improbable points can be handled with a restriction of $\mathcal{X}$. $C_I$ also misses some points of $B_I$, if a point is associated with the baseline target value, there still may be other points with the same projection on $I$ and with different labels. Thus, 
\begin{center}
\begin{minipage}{0.475\linewidth}
    \begin{equation}
        \label{c_dual}
        A_I \subset \overline{C_{\bar{I}}}
    \end{equation}
\end{minipage}
\hfill
\begin{minipage}{0.475\linewidth}
    \begin{equation}
       \label{g_dual}
        A_I \subset \overline{G_{\bar{I}}}
    \end{equation}
\end{minipage}
\end{center}
Gradient-based and baseline-comparison proxies are \textbf{linked to the formalisation we derive and provide upper bounds for functionality domains}. In section \ref{sec:exp} we quantify how good these two approximations are.

\subsection{Selector-predictors}

A final recent idea is to try to incorporate and learn the instance-wise selection task during training \cite{chen2018learning, yoon2018invase, arik2019tabnet, yamada2020feature}. These techniques have been referred to as \textit{selector-predictor} \cite{camburu2020explaining}. The idea is to use two models: a \textit{selector} $\textrm{Sel}: \mathcal{X} \mapsto \{ 0,1 \}^n$ whose goal is to determine a map $S$ of the most-relevant features for each point; and a \textit{predictor} $\textrm{Pred}: \mathcal{X} \mapsto \mathcal{Y}$ acting as the usual prediction model of $Y$ with the twist that it takes $X \odot S$ as input. The training objective varies between methods but the general spirit is to maximise the performances of $\textrm{Pred}(X \odot \textrm{Sel}(X))$ at predicting $Y$ while either minimising the number of selected features in $\textrm{Sel}(X)$ or ensuring the constraint that $k < n$ features are selected.
A first issue is that most of these methods are only evaluated on performance-degradation metrics or on rather global synthetic selection tasks, which do not truly evaluate instance-wise interpretations. A second, more alarming, issue is that the selector model is completely free and prone to degenerate selection solutions (see Supplementary B). 
In particular, the selector does not verify properties \ref{prop:hyperspace} and \ref{prop:parents}.

\subsection{Relational database connections}

We should lastly mention that we found our formalisation to resemble the concept of \textit{functional dependency} from relational database theory \cite{armstrong1974dependency}. Our simple categorical attribution \eqref{measure:prob} is strikingly similar to \cite{kivinen1995approximate}. But the purpose is not interpretation and in this latter field, global multi-dependence among all columns of a table are sought, differently from between a subset of the input and a designated output, and, to our knowledge, not in an instance-wise manner.


\section{Experiments}
\label{sec:exp}

Armed with a formalism, we generate synthetic distributions with instance-wise ground-truth selections to evaluate attributions methods approximate selection performances and check their solution structure.
All generated data, implementations and evaluations methods are available and fully reproducible at our paper code repository~\footnote{Source code at \href{https://github.com/deezer/functional_attribution}{github.com/deezer/functional\_attribution}}.

\subsection{Synthetic tasks with ground-truth selections}

In this section we first explain how, from a desired selection random variable $S^*$, we are able to build a distribution $p_{X,Y}$ with a given selection solution $S^*$, \textit{i.e.}
\begin{equation*}
    S^* = \arg\min\limits_{I \subset [n]} X \in A_I(p_{Y\mid X})
\end{equation*}
note that $A_I$ depends on $p_X$. As most selection methods do not handle multiple minimal solution well, we restrict our study to the case with unique selection minimum.

\begin{figure}[t]
  \centering
  \includegraphics[trim={3cm 1cm 1cm 1.5cm},clip,width=0.9\linewidth]{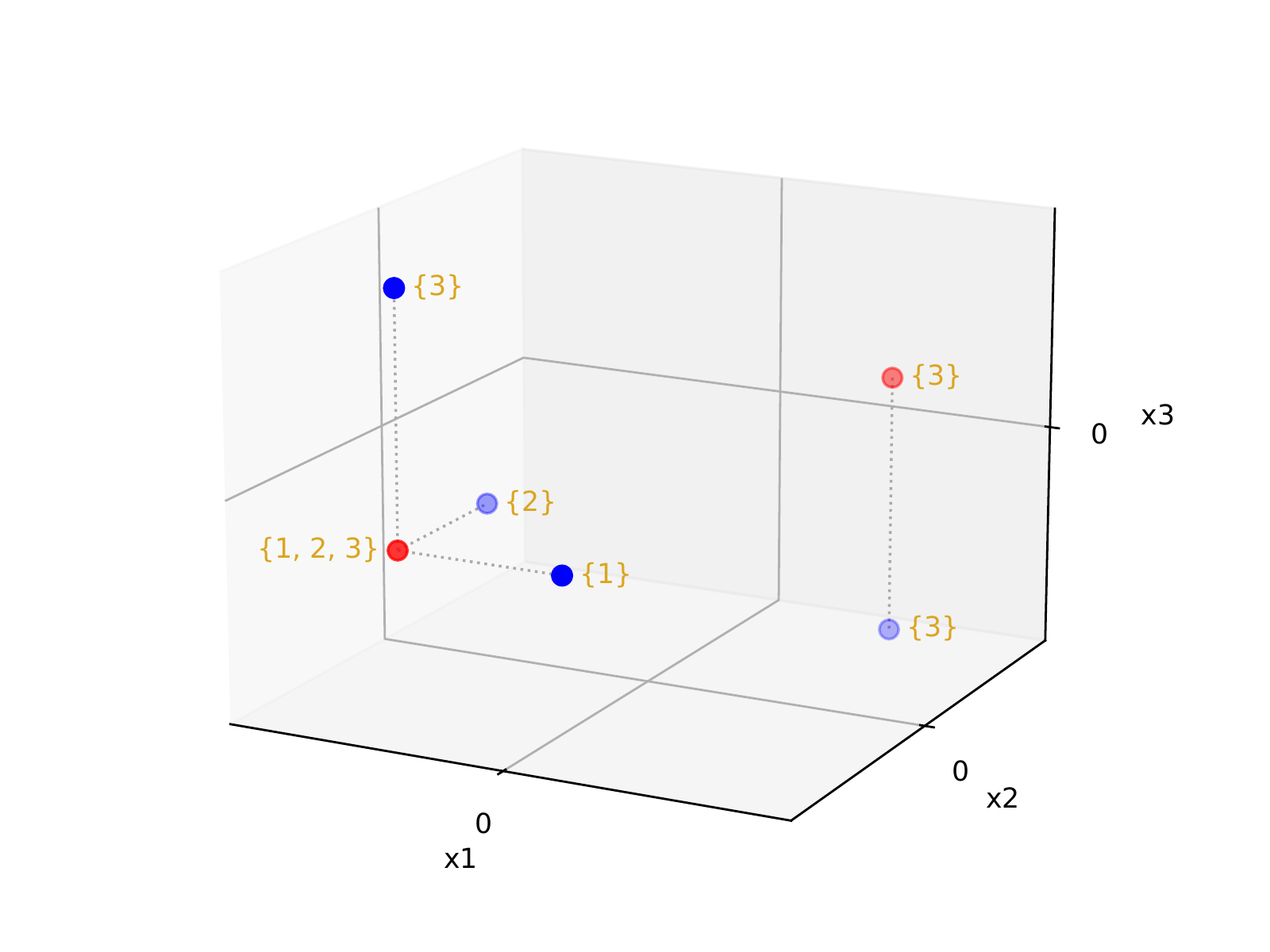}
  \caption{Example of generated distribution $p_{X,Y}$ from a list of six centroids in $\mathbb{R}^3$, with associated labels in blue (0) and red (1), and corresponding unique selection solution $S^*(x)$ in yellow. This example can be found in our dataset under the name \texttt{task 3\_19}.}
  \label{fig:3_19}
\end{figure}

We consider the following simple generative process to draw the data: we uniformly sample from a finite list of points $(c_1, ... c_m) \in \mathcal{X}$ -- we call \textit{centroids} -- with an associated binary label $y_j$ in $\mathcal{Y} = \{ 0, 1 \}$. This is our \textbf{perfect-dependence} distribution $p_{X,Y}$:
\begin{align*}
& C \sim \mathcal{U}\{1, ... m \} \\
& \mathbb{P}(X = c_j, Y = y_j) = \mathbb{P}(C = j)
\end{align*}
As we are in a binary case, interpreting $p_{Y\mid X}$ can be reduced to the study of the optimal mapping $f = \mathbb{P}(Y=1 | X=x)$. Since we want to assign a unique selection subset $S^*(x) \in [n]$ to each point, we need to ensure that $S^*(x)$ is indeed the minimal subset such that $x \in A_{S^*(x)}(f)$. 
To do that, we choose the centroids in order to have neighbors with opposite labels in each direction of $S^*(c_j)$ exclusively, so that we know that $c_j \in A_{S^*(c_j)}(f)$, and that for all $J \subset [n]$ such that $J \cap S^*(c_j) \neq \emptyset$, we have $c_j \in B_{J}(f)$. An example is shown in figure \ref{fig:3_19}.

To have a continuous distribution and allow gradient computations, we then replace our discrete points with normal distributions with fixed variance $\sigma^2$. We obtain a familiar Gaussian mixture distribution $p'_{X,Y}$:
\begin{align*}
    p'_{X|C}(x \mid c_j) &= \mathcal{N}(x ; c_j, \sigma^2) \\
    p'_{X,Y}(x,y) &= \sum\limits_{j=1}^m p_C(c_j)p'_{X|C}(x \mid c_j)\delta_{y = y_j}
\end{align*}
The dependencies are now \textbf{imperfect}, we evaluate the capacities of attribution methods to return $S^*$ given $p'_{X,Y}$ and the imperfect optimal mapping $f' = p'_{Y\mid X}(y=1 \mid x)$.

With this principle, we are able to generate synthetic distributions of any dimension, with unique selection ground-truth of any dimension. The full generation algorithm, more details and examples are given in the Supplementary C. 

\subsection{Considered methods}

With no requirement to learn a mapping from $X$ to $Y$, or to make prior assumption on the selection space, many methods collapse into one. Additive mixture of experts methods can all be summarised as the evaluation of \textit{GAM}, \textit{GA\textsuperscript{2}M} with added pairwise interactions, ... up to \textit{GA\textsuperscript{$\infty$}M} that considers all possible input subset restrictions, thus with an exponential complexity.
Note that \textit{GA\textsuperscript{$\infty$}M} is equivalent to the weighted \textit{fANOVA} without the \textit{integration-to-zero} that hampered instance-wise selection. 
Generalised additive models do not directly define attribution values, but since their finality is to estimate $\mathbb{E}[Y|X_I]$, we can use the relaxation \eqref{measure:prob} we derived in section \ref{sec:relaxation}. We thus dub them with a \textit{"attr"} prefix to underline the modification of the original models. 
We analyse two supplementary recent methods that we deemed sufficiently different from generalised additive models: \textit{InterpretableNN} \cite{afchar2020making}, based on \textit{GA\textsuperscript{$\infty$}M} with a custom selection mechanism inspired by boosting; and \textit{Archipelago} \cite{tsang2020does}, based on \textit{GA\textsuperscript{2}M}, that merges found pairwise dependence using a union-find algorithm to yield disjoint subset selection candidate with a quadratic complexity. Among proxy methods, we evaluate \textit{LIME} \cite{ribeiro2016should} in both categorical (\textit{Cat.}) and continuous (\textit{Cont.}) configurations;
all gradient-based methods cited in \ref{sec:proxy_methods}; the sampled classic shapley value estimation \cite{vstrumbelj2014explaining}  -- $\mathbb{E}(f')$, and the baseline approximation introduced in SHAP \cite{NIPS2017_7062} -- $f'(\mathbb{E})$.
The selector-predictors are the only methods for which we have to sample from $p'_{X,Y}$ and train two neural networks, we evaluate \textit{L2X} \cite{chen2018learning} with a fixed number of sampled selection dimensions, and \textit{INVASE} \cite{yoon2018invase} that notably replaces this constraint with a Lagrangian penalty in its objective.

\subsection{Methods evaluation}

We generate \textbf{1000 supervised tasks with ground-truth unique univariate selections} -- $S^*(c_j)$ is a singleton for all centroids;
and \textbf{1000 tasks with unique multivariate selections} -- $S^*(c_j)$ has a cardinality $k(c_j)$ and is chosen among $n \choose k(c_j)$ possible subsets. We additionally generate 100 multivariate tasks to tune $\eta$ for each method. The input space dimension is gradually raised from $\mathbb{R}^2$ to $\mathbb{R}^{11}$, leading up to $2^{11}$ possible selection subset candidates per centroid.

Our results for univariate selection are given in table \ref{table:univariate}. \textit{Archipelago}, \textit{InterpretableNN} and \textit{attr-GA\textsuperscript{k}M} methods are all equivalent when returning univariate solutions. We use the standard accuracy metric between the predicted $\hat{S}$ and ground-truth selection $S^*$ on each centroid. Only generalised additive models equipped with the attribution measure \eqref{measure:prob} and shapley-based methods solve the tasks perfectly. We note that this latter method counter-part, \textit{SHAP}, with the baseline choice $f'(X_{I}, \mathbb{E}_{\bar{I}}(X_{\bar{I}} ) )$ especially underperforms despite its complexity. This had already been noticed \cite{slack2020fooling} and is due to the fact that the baseline requires out-of-distribution evaluations of $f'$. For fairness, we also include a performance evaluation (\textit{Acc\textsuperscript{*}}) leveraging the prior knowledge that the ground-truth solutions are singletons -- \textit{i.e.} selecting the singleton of maximum responsibility.

\begin{table}[h]
\small
\centering
\caption{\textbf{Feature selection performance on 1000 univariate tasks} of attributions methods under study, with 95\% confidence interval indicators and total computation time $T$. }
\vskip 0.1in
\begin{tabular}{c|c|c|c}
    \toprule
    Method & Acc (\%) & Acc\textsuperscript{*} (\%) & $T$ (h:m:s) \\
    \midrule
    \textit{LIME} (Cat.) & 32.4 $\pm$ 1.8 & 61.9 $\pm$ 0.8 & 0:00:54 \\
    \textit{LIME} (Cont.) & 10.6 $\pm$ 1.0 & 43.1 $\pm$ 0.9 & 0:00:47 \\
    \textbf{\textit{attr-GAM}} & \textbf{100} & \textbf{100} & 0:00:10 \\
    \textbf{Shapley} ($\mathbb{E}(f)$) & \textbf{100} & \textbf{100} & 0:05:36 \\
    \textit{SHAP} ($f(\mathbb{E})$) & 23.1 $\pm$ 1.2 & 37.9 $\pm$ 1.2 & 0:05:39 \\
    Gradient & 33.5 $\pm$ 1.0 & 87.8 $\pm$ 0.9 & 0:00:02 \\
    Grad$\times$Input & 32.2 $\pm$ 1.0 & 88.4 $\pm$ 0.9 & 0:00:02 \\
    \textit{Integrated Grad.}  & 38.5 $\pm$ 0.9 & 80.6 $\pm$ 1.2 & 0:00:05 \\
    \textit{Expected Grad.} & 45.8 $\pm$ 0.9 & 63.3 $\pm$ 0.8 & 0:00:20 \\
    \midrule
    \textbf{\textit{attr-GA\textsuperscript{$\infty$}M}} & \textbf{100} & \textbf{100} & 0:01:10 \\
    \midrule
    \textit{L2X} & 51.8 $\pm$ 1.1 & 52.5 $\pm$ 1.0 & 19:33:37 \\
    \textit{INVASE} &  26.5 $\pm$ 1.1 & 35.5 $\pm$ 1.1 & 45:36:49 \\
    \bottomrule
\end{tabular}
\vskip -0.1in
\label{table:univariate}
\end{table}

In table \ref{table:multivariate}, we show the results for selection tasks with ground-truth selections subsets of any cardinality, which is particularly more difficult. The best performing models are still the generalised additive-based and shapley-value-based models. It must be noted that, with synthetic distributions, all methods have access in $O(1)$ to $p'(Y=1\mid X_I=x_I)$ for all subset $I$, whereas we have to let selector-predictors methods learn it from scratch to properly evaluate their selector, hence their high computation time $T$. Additive model methods are all derived from \textit{GA\textsuperscript{$\infty$}M} and use caching for faster inferences, we thus only display order of magnitude for $T$. \textit{INVASE} particularly underperforms, we believe that this may be magnified by the difficult tuning of its sparsity-inducing penalty term (see Supplementary D). 

\begin{table}[h]
\small
\centering
\caption{\textbf{Feature selection performance on 1000 multivariate tasks} for attributions methods under study. }
\vskip 0.1in
\begin{tabular}{c|c|c}
    \toprule
    Method & Acc (\%) & $T$ (h:m:s) \\
    \midrule
    \textit{LIME} (Cat.) & 16.2 $\pm$ 1.3 & 0:05:54 \\
    \textit{LIME} (Cont.) & 27.4 $\pm$ 1.6 & 0:05:47 \\
    \textit{attr-GAM} & 24.5 $\pm$ 1.5 & 0:00:25 \\
    Shapley ($\mathbb{E}(f)$) & 74.3 $\pm$ 1.1 & 0:16:29 \\
    \textit{SHAP} ($f(\mathbb{E})$) & 15.7 $\pm$ 1.3 & 0:17:41 \\
    Gradient & 26.5 $\pm$ 1.5 & 0:00:04 \\
    Gradient$\times$Input & 22.6 $\pm$ 1.5 &  0:00:04 \\
    \textit{Integrated Gradient} & 18.5 $\pm$ 1.4 & 0:00:24 \\
    \textit{Expected Gradient} & 21.4 $\pm$ 1.4 & 0:03:42 \\
    \midrule
    \textbf{\textit{attr-GA\textsuperscript{$\infty$}M}} & \textbf{81.7 $\pm$ 1.1} & 0:17:44\textsuperscript{$*$} \\
    \textit{attr-GA\textsuperscript{2}M} & 52.5 $\pm$ 1.8 & $\ll *$ \\
    \textit{attr-GA\textsuperscript{3}M} & 74.1 $\pm$ 1.3 & $< *$ \\
    \textit{attr-GA\textsuperscript{4}M} & 81.2 $\pm$ 1.1 & $< *$ \\
    \textit{InterpretableNN} & 79.7 $\pm$ 1.2 & $\simeq *$ \\
    \textit{Archipelago} & 70.2 $\pm$ 1.1 & $\simeq *$ \\
    \midrule
    \textit{L2X} & 23.7 $\pm$ 1.6 & 32:53:16 \\
    \textit{INVASE} & 7.4 $\pm$ 0.9 & 44:15:44 \\
    \bottomrule
\end{tabular}
\vskip -0.1in
\label{table:multivariate}
\end{table}

\subsection{Necessary property evaluation}

We finally link the performance differences we observe, to the necessary properties we derived in section \ref{sec:necessary} and check whether the methods provide well structured selection solutions. Using the predicted selections of the multivariate tasks, we compute the ratio of centroids verifying property \ref{prop:hyperspace}. To do that, we leverage property \ref{prop:parents}: \ref{prop:hyperspace} is verified for a centroid $c_j$ \textit{iff} for a given selection $\hat{S}(c_j)$, for every centroids $c_k$ such that $P_{\hat{S}(c_j)}(c_j) = P_{\hat{S}(c_j)}(c_k)$, we have $\hat{S}(c_k) \subseteq \hat{S}(c_j)$. 

The results are presented in table \ref{table:propverification}. Strikingly, we observe a correlation between the property-verification and feature selection performances ($\rho = 0.88$). This is a strong indication that \textbf{well structured selections solutions} with regards to all points in the input distribution \textbf{tend to also be better performing}. Interestingly, we must underline that computing the property verification rate \textbf{does not require to have access to ground-truth selections}, which opens the door for further applications to real-data tasks. We must however emphasize that property \ref{prop:hyperspace} is not sufficient: \textit{e.g.} globally returning all or no input features for all points as selection is a perfectly structured solution according to \ref{prop:hyperspace}, but rarely an optimal one. 
Property \ref{prop:hyperspace} is necessary but not sufficient. It must help design better instance-wise attribution methods that intrinsically verify it, as \textit{GA\textsuperscript{$\infty$}M}, but should not be a criterion to maximise.

\begin{table}[h]
\small
\centering
\caption{\textbf{Ratio of points verifying property \ref{prop:hyperspace} on 1000 multivariate tasks}. Note that this does not require any ground-truth label, only the proposed selection solution is analysed. }
\vskip 0.1in
\begin{tabular}{c|c}
    \toprule
    Method & Property verification rate (\%) \\
    \midrule
    \textit{LIME} (Cat.) & 29.9 $\pm$ 1.7 \\
    \textit{LIME} (Cont.) & 46.6 $\pm$ 1.6 \\
    \textit{attr-GAM} & 61.5 $\pm$ 1.1 \\
    Shapley ($\mathbb{E}(f)$) & 79.5 $\pm$ 1.1 \\
    \textit{SHAP} ($f(\mathbb{E})$) & 23.7 $\pm$ 1.5 \\
    Gradient & 61.6 $\pm$ 1.3 \\
    Gradient $\times$ Input & 54.5 $\pm$ 1.3 \\
    \textit{Integrated Gradient} & 39.7 $\pm$ 1.5 \\
    \textit{Expected Gradient} & 41.8 $\pm$ 1.5 \\
    \midrule
    \textbf{\textit{attr-GA\textsuperscript{$\infty$}M}} & \textbf{92.9 $\pm$ 0.6} \\
    \textit{attr-GA\textsuperscript{2}M} & 63.7 $\pm$ 1.4 \\
    \textit{attr-GA\textsuperscript{3}M} & 81.2 $\pm$ 1.4 \\
    \textit{attr-GA\textsuperscript{4}M} & 90.7 $\pm$ 1.1 \\
    \textit{InterpretableNN} & 86.9 $\pm$ 0.9 \\
    \textit{Archipelago} & 88.8 $\pm$ 0.7 \\
    \midrule
    \textit{L2X} & 37.5 $\pm$ 1.6 \\
    \textit{INVASE} &  61.3 $\pm$ 1.7 \\
    \bottomrule
\end{tabular}
\vskip -0.1in
\label{table:propverification}
\end{table}


\section{Conclusion}

The growing interest in \textit{interpretable machine learning} and the profusion of recent feature attribution methods has motivated us to take a step back and propose a rigorous formalisation of often vaguely defined concepts in this field. Though to some extent task-dependent, we argue that all these methods can be analysed through an irreducible component: feature selection.
Doing so, we could evaluate many state-of-the-art methods on rigorously derived ground-truth rationales, and we have derived provable necessary properties that any computed interpretations must verify -- which is not the case for some popular methods.
Our future directions involve using our relaxation framework to derive good attribution measures for specific applications and building new efficient and well-formulated attribution models.

\clearpage
\bibliography{biblio.bib}
\bibliographystyle{apalike}

\appendix

\section*{Supplementary Material}

We list some general notations we use throughout this paper:

\begin{tabularx}{\linewidth}{ l|X }
\toprule
Symbol & Meaning \\
\midrule
$\wedge$ & Logical \texttt{AND} \\
$\mathcal{X}, \mathcal{Y}$ & input and output space \\
$X, Y$ & input and target random variable (r.v.) \\
$x, y$ & input and target sample \\
$p_X, p_{Y\mid X}$ & distribution of $X$; $Y$ conditional to $X$ \\
$\mathcal{U}$ & uniform distribution \\
$\mathcal{N}$ & multivariate normal distribution \\
$\mathcal{B}$ & Bernoulli distribution \\
$\vec{e}_i$ & i-th canonical vector of $\mathbb{R}^n$ \\
$[n]$ & set of integer from 1 to $n$ \\
$I$ & subset of integer \\
$f$, $f_i$, ... & denotes a function \\
$R$, $R_i$, ... & denotes a binary relation (b.r.) \\
$D$ &  denotes a dataset (hence defines a b.r.) \\
$X_I$ & r.v. $X$ projected to the input features with indexes $I$ \\
$R_I$ & b.r. $R$ with its domain projected to $I$ \\
$f_I$ & function with its domain projected to $I$ \\
$\mathbb{E}_{Y\mid X}$ & mean equipped with the distribution of $Y\mid X = x$ \\
$\delta_A$ & indicator function of set $A$ \\
\bottomrule
\end{tabularx}

\section{fANOVA instance-wise selection failure}
\label{annex:anova}

We detail our claim that the fANOVA is indicative of a global feature dependence, but not an instance-wise one, with a simple example.

Let us take a binary bidimensional problem with input $X = (X_1, X_2) \in \{0,1\}^2$ following a uniform probability and we try to compute the subset dependence of a \texttt{AND} function $f$ -- \textit{i.e.} $Y = f(X) = X_1 \wedge X_2$. The fANOVA gives the unique decomposition:
\begin{align*}
    f_{\emptyset} &= 1/4 = \mu  &= \begin{bmatrix} \mu & \mu \\
                           \mu & \mu
                        \end{bmatrix} \\
    f_1 &= \begin{bmatrix} -\mu & \mu \end{bmatrix}
                        &= \begin{bmatrix} -\mu & \mu \\
                           -\mu & \mu
                        \end{bmatrix} \\
    f_2 &= \begin{bmatrix} -\mu \\ \mu \end{bmatrix}
                        &= \begin{bmatrix} -\mu & -\mu \\
                           \mu & \mu
                        \end{bmatrix} \\
    f_{1,2} &= \begin{bmatrix} \mu & -\mu \\
                              -\mu & \mu
                        \end{bmatrix} &
\end{align*}
For convenience, we have represented the functions as matrices indicating their four possible values for $X$, where $X_1$ varies along the columns and $X_2$ along the rows. We check that the value given by $f_1$ (\textit{resp.} $f_2$) only depends on $X_1$ (\textit{resp.} $X_2$). Then we check the identifiability constraint that each function other than $f_{\emptyset}$ is zero-centered, and that we indeed obtain a decomposition of $f$:
\begin{align*}
    f & = f_{\emptyset} + f_1 + f_2 + f_{1,2} \\
     & = \begin{bmatrix} 0 & 0 \\
                         0 & 4\mu
        \end{bmatrix} \\
             & = \begin{bmatrix} 0 & 0 \\
                         0 &1
        \end{bmatrix} \\
    & = X_1 \wedge X_2
\end{align*}

Meanwhile, the selection solution is the following:
\begin{align*}
    \textrm{attr}(f) &= \begin{bmatrix} \{ \{1\}, \{ 2 \} \} &
                                    \{ \{1\} \} \\
                                    \{ \{2\} \}  & \{ \{1, 2\} \} \\
        \end{bmatrix}
\end{align*}
which can be found by testing all four possible restrictions. For instance, for $p_0 = (0,0)$, we have $p_0 \in A_1(f)$ and $p_0 \in A_2(f)$ as the value of $f$ is constant and equal to zero in their respective complementary directions, and those two solutions are minimal -- \textit{i.e.} $p_0 \not\in A_\emptyset(f)$ as we have a different 1 value on the complementary direction $\overrightarrow{(1,1)}$.

This solution is not translated from the strong symmetry of the ANOVA. For instance, the bivariate term only appear when $X = (1,1)$, but $f_{1,2}$ is non-null everywhere. The same discussion applies for $X = (0,0)$ for which two solutions co-exist, but this is not obvious only looking at the ANOVA decomposition. As mentioned, this is due to the centering constraint that creates "artifacts" in all subfunctions -- we say this from the standpoint of instance-wise attribution.

\section{Selector-predictor degenerate selection solutions}
\label{annex:selpred}

We show with a constructive counter-example that selector-predictor methods have degenerate solutions that are optimal relative to their objective function, but with meaningless selected features. Our main point will be that, contrary to the intuition behind selector-predictor, splitting the model into a selector and a predictor does not enable to split the joint prediction and selection task between the two: predictions abilities may percolate to the selector and vice-versa. We show an extreme solution where the full prediction work is done by the selector.

We study the model \textit{L2X} \cite{chen2018learning} for which the selector uses a reparametrisation trick to sample $k$ selections of features. We do not fully detail the method since most of the difficult labor is related to finding a relevant relaxation allowing to train the selector model $\textrm{Sel}$ with its discretised output through a gradient-descent. We skip training and directly study theoretical optimal parameters. The objective function of the model on each target sample $(x,y)$ is the cross-entropy loss $l\left( y ; \textrm{Pred}(x \odot v(x))  \right)$, where $v$ is a feature mask sampled for each $x$ with a predefined and fixed number $k$ of non-null values.

During training, sampled $v$ are relaxed to $[0,1]^n$ and have to explore different values for a given $x$ around logit predictions of the selector $\textrm{Sel}(x)$. Again, we ignore these details as after the training phase $\textrm{Sel}$ is trained and frozen, $v$ is not longer sampled, it is binary and deterministically mapped from $x$ as the top-$k$ logit values of $\textrm{Sel}(x)$. For simplicity, we absorb the top-$k$ filtering into $\textrm{Sel}$ and directly denote with $\textrm{Sel}(x)$ the binary mask of the $k$ selected features.
For our selector and predictor model parametrised by $\theta$, we will thus write the expected loss after training $\mathcal{L}(\theta) = \mathbb{E}[l(Y ; \textrm{Pred}_\theta (X \odot \textrm{Sel}_\theta(X) ) )]$. We denote the minimal theoretical loss $\hat{\mathcal{L}} = \min\limits_\theta \mathcal{L}(\theta)$.

We can compare this loss with the \textit{unmasked} case where we would simply predict $Y$ without restricting the number of usable input features in the predictor: $\hat{\mathcal{L}}_u = \min\limits_\theta \mathbb{E}[l(Y ; \textrm{Pred}_\theta (X)) ]$. In general, we have $\hat{\mathcal{L}} \geq \hat{\mathcal{L}}_u$ as the expressive power of the predictor is superior when having access to any order of interaction between variables.

Here, we assume that our task admits a unique instance-wise selection solution everywhere with exactly $k$ features, \textit{i.e.} the parameter $k$ in the selector is well tuned and the problem well-posed. This means that given the ground-truth selection random variable $S^*$, we have that $\mathcal{L}^* = \min\limits_\theta \mathbb{E}[l(Y ; \textrm{Pred}_\theta (X \odot S^*) )]$ will be equal to the optimal loss in the unmasked case $\hat{\mathcal{L}}_u$ as $S^*$ only captures features that are relevant to predict $Y$.
Now, if the selector could approximate $S^*$ with a set of parameters $\theta^*$, we would thus have $\mathcal{L}(\theta^*)$ tends to $\mathcal{L}^*$, and consequently $\mathcal{L}(\theta^*)$ tends to  $\hat{\mathcal{L}}_u$.
However, we do not have access to $S^*$ and cannot evaluate how good the approximation is, we can only compare $\mathcal{L}(\theta)$ to its theoretical bound $\mathcal{L}^*$, that, when we assume that $k$ is well chosen, is equal to the observable unmasked bound $\hat{\mathcal{L}}_u$. The question we should now ask ourselves is whether having found parameters $\hat{\theta}$ such that $\mathcal{L}(\hat\theta) = \hat{\mathcal{L}} = \hat{\mathcal{L}}_u = \mathcal{L}^*$ means that we have $\textrm{Sel}_{\hat{\theta}}(X) = S^*$?

This is critical as selector-predictor models are evaluated in comparison to their non-input-restricted counterpart as a proxy of their approximation of $S^*$: it is often shown that there is no significant drop in performances while selecting a minimal number of input features. We now show that there exists many equivalent and optimal solutions $\theta'$ such that $\mathcal{L}(\theta') = \mathcal{L}^*$ and $\textrm{Sel}_{\theta'}(X)$ verifies the constraint of selecting $k$ features but while being nowhere close to $S^*$ or to having any interpretation value.

For that we consider the case of a categorical task, and assume that $Y$ is one-hot encoded as $g(Y) \in \{ 0, 1 \}^C$ onto the $C$ possible classes in $\mathcal{Y}$, where $g$ denotes the one-hot-encoding function. Additionally, we introduce a random permutation $\sigma$ of $[C]$, and its inverse permutation $\sigma^{-1}$. We assume we know the ground-truth value of $k$. Finally, we assume that the input dimension $n$ is greater than $C + k - 1$, which is quite common, and we denote a padding operator $p: \mathbb{R}^C \mapsto \mathbb{R}^n$ that completes any vector of size $C$ with $k-1$ ones and $n - (k-1)$ zeros to fit in $\mathbb{R}^n$.

Now, as in \textit{L2X}, we assume that the predictor and selector are parametrised with two families of neural networks with comparable number of parameters. We denote $f_\theta$ a member of the selector neural network family and delete $n - C$ neurons in the output layer to obtain a $C$-dimensional output. Though $f_\theta$ belongs to the selector family, we use it to approximate $Y$: 
we denote by $\hat{\theta}_f$ some optimal parameter associated with the optimal loss
$\hat{\mathcal{L}}_s = \min\limits_\theta \mathbb{E}[l(g(Y); f_\theta (X))]$\footnote{We assume that $f_\theta$ outputs probability vectors, \textit{e.g.} using a  \texttt{softmax} in its last layer. Before this expression the one-hot encoding operations $g(Y)$ were eluded in the losses.}. And we have,
\begin{equation*}
    \hat{\mathcal{L}}_s \simeq \hat{\mathcal{L}}_u
\end{equation*}
The $\simeq$ holds if the selector and predictor families have a comparable expressive power. We have an equality with the default implementation of \textit{L2X}.

Then, we come back to the selector-predictor objective and the trick is to study the solution given by
\begin{align*}
    \textrm{Sel}_\sigma(x) &= p(\sigma(f_{\hat{\theta}_f}(x))) \\
    \textrm{Pred}_\sigma(x) &= \sigma^{-1} \left( \begin{bmatrix} \delta_{|x_1| > 0} \\ \vdots \\ \delta_{|x_C| > 0}  \end{bmatrix}
    \right)
\end{align*}
We check that this solution is indeed part of the parametrised family for the predictor and selector. We have built $f_\theta$ to have the right selector architecture, except for $n-C$ missing neurons in its last layer; the composed permutation $\sigma$ can be crafted by permuting the $C$ output neurons of $f_\theta$; composing by $p$ is done by adding $n - C$ constant neurons in the output layer and we thus obtain the right architecture. As for the predictor, the only new element is the non-null indicator functions on the $C$ first input feature. If we were to assume the activation were the step function $H$, this would be straight-forward to approximate with three neurons, \textit{e.g.}
\begin{equation*}
    \delta_{|x_i| > \epsilon} = H(H(x_i \geq \epsilon) + H(- x_i \geq -\epsilon) \geq 2)
\end{equation*}
With \textit{sigmoid} and \textit{ReLU} activations, this can be done using big multiplicative coefficients. Overall, we need $\mathcal{O}(C)$ neurons on two layers to approximate the function $\textrm{Pred}_\sigma$. The other $n-C$ input features of the predictor are ignored using zero weights in the neurons parameters.
It is reasonable to think that with neural network architectures used in practice, $\textrm{Pred}_\sigma$ is indeed part of the predictor parametrised family, or can be closely approximated.

We denote the found parameters $\hat{\theta}_\sigma$. This particular solution enables us to have
\begin{gather*}
    \textrm{Pred}_{\hat{\theta}_\sigma}(x \odot \textrm{Sel}_{\hat{\theta}\sigma}(x) ) = f_{\hat{\theta}_f}(x) \\
    \mathcal{L}({\hat{\theta}_\sigma}) \simeq  \hat{\mathcal{L}}_u
\end{gather*}

In essence, \textbf{we are estimating $Y$ in the selector and encoding this information in the selection mask we pass to the predictor}. We check that the found selector returns a binary mask with exactly $k$ non-null components: one in the $C$ first components that encodes the label, $k-1$ in the padding operator for the remaining features.

What about the found selections? We have $C!$ possible optimal set of parameters $\hat{\theta}_\sigma$, all of them maximal according to the \textit{L2X} objective function, with them, all first $C$ features of $X$ can be made equally maximally important for selection, regardless of data. We conclude that the selector solution does not translate any truth about dependence between $Y$ and $X$. A even more efficient label-passing degenerate case can be obtained by replacing $g$ with a function that encodes labels with binary numbers, only requiring $n > \log_2(C) + k - 1$ instead of $n > C + k - 1$.

\textit{INVASE} \cite{yoon2018invase} is similar to \textit{L2X} with a Lagrangian penalty instead of constraining to have exactly $k$ non-null selected features; it is similarly prone to degenerate selection solutions, but it goes even further. It must be noticed that we only require to output one non-null component in the selector to pass the true label and have an optimal prediction. This means that with a ground-truth selection cardinality $k > 1$, \textbf{our degenerate solutions yield an optimal prediction loss with a lower regularisation penalty than when using $\textbf{S}^*$}, since $S^*$ may have more than one required features for selection. We have observed such effect in practice: \textit{INVASE} has good prediction performances and returns very sparse selection masks correlated with ground-truth labels and having nothing to do with ground-truth selection.

One way to avoid label-passing issues is to verify the properties 1 and 2 we propose.

\section{Tasks generation}
\label{annex:task}
 
In this section we explain in detail how the centroids $(c_1, ... c_m)$, their label $y_j$ and ground-truth selection $s^*_j$ are chosen. The unifying condition of these latter variables is that $s^*_j \subset [n]$ should be the unique subset of minimal cardinality verifying $c_j \in A_{s^*_j}(f)$, with $f = p(y=1|x)$.

\subsection{Binary Hypercube}

\begin{figure}[h]
  \centering
  \includegraphics[trim={0 0 1cm 1cm},clip,width=\linewidth]{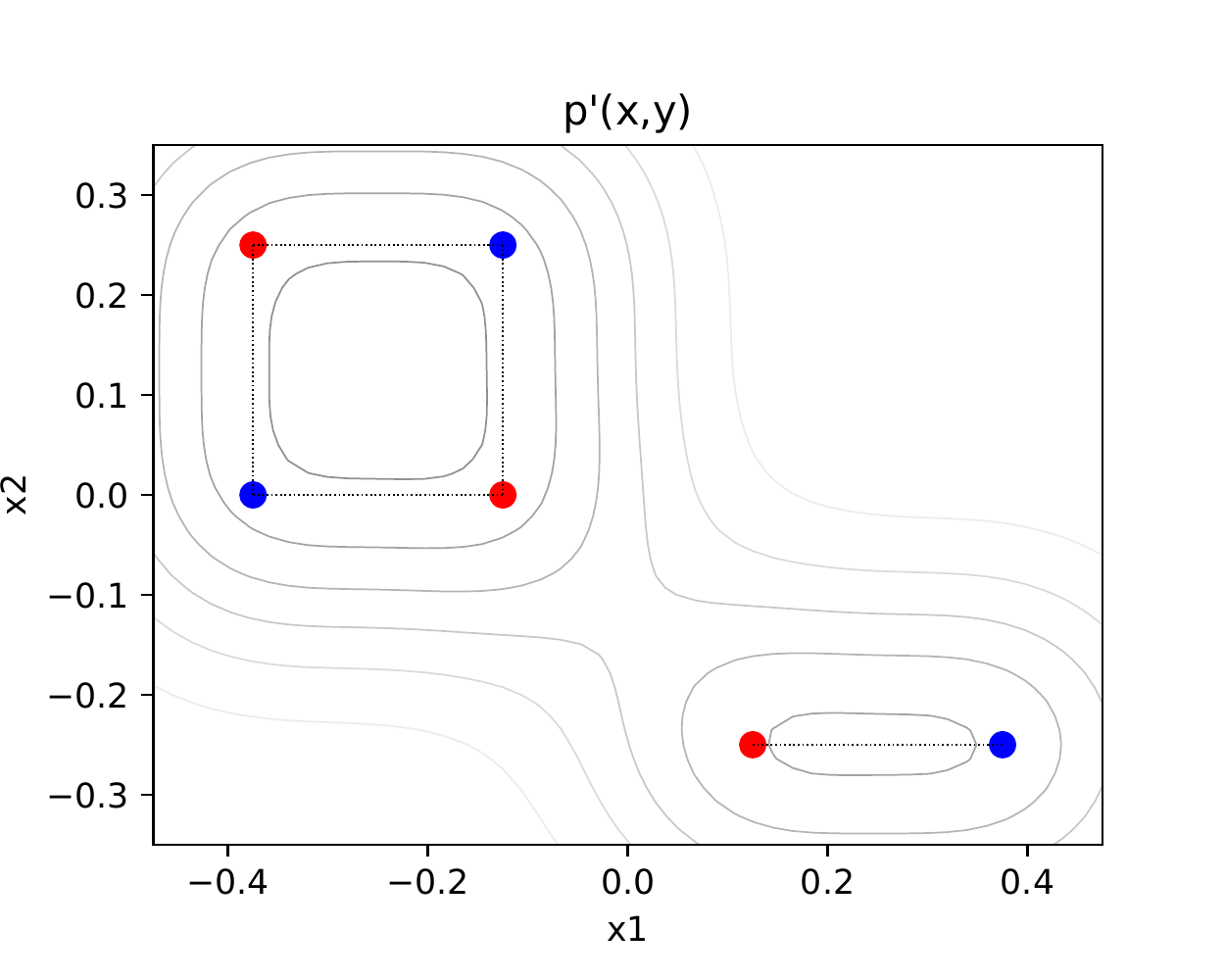}
  \caption{\textbf{Distribution of task \texttt{2\_14}} The centroid $c_j$ are indicated with colored dots -- red for $y_j = 1$ and blue for $y_j = 0$. Dotted lines connect centroids with opposite labels and that differ by only one coordinate, \textit{i.e.} that will be superposed if projected on the other coordinate. The corresponding Gaussian mixture for the distribution with imperfect dependence is hinted in black contours.} 
  \label{fig:2_14_a}
\end{figure}

We first propose to study the case of centroids forming the vertices of an hypercube. For that, we choose a subset of indexes $J^k \subset [n]$ and study a set $Q^k$ that contains the vertices coordinates of an hypercube of dimension $|J^k|$ placed in $\mathcal{X}$ with its edges aligned with the canonical vectors $\{ \vec{e_i} \mid i \in J^k \}$.
Since hypercube graphs are bipartite \cite{foldes1977characterization}, we can assign a binary label to each vertex with the nice property that for a given point $x \in Q^k$ and associated label $y$, each single coordinate change to $x$ to find another point $x' \in Q^k$ will yield a neighbor associated to an opposite label $\bar{y}$ (\textit{i.e.} we can color the graph with labels $y$ and $\bar{y}$). This is illustrated in figure \ref{fig:2_14_a}, we display a generated distribution $p_{X,Y}$ with centroids defined using two hypercubes: one of dimension 2 (also known as the \textit{XOR} problem) and another of dimension 1 oriented along $\vec{e_1}$. We have added dotted lines to highlight the edges of the considered hypercubes.
Therefore, for all $x \in Q^k$, for all $i \in J^k$ we have $x \in B_i(f)$: all points of $Q^k$ have neighbors with \textit{contradicting labels} along the dimensions indexed in $J^k$. Using the property 2 on hierarchy, for $H \subset [n]$ such that $H \cap J^k \neq \emptyset$, \textit{i.e.} if $H$ contains at least one index of $J^k$, we have $x \in B_H(f)$. By defining $H' \subsetneq J^k$ and choosing 
$H = [n] \setminus H'$, we check that $H \cap J^k =  J^k \setminus H' \neq \emptyset$, and we obtain $x \in B_H(f)$ thus $x \not\in A_{H'}(f)$ for all $H' \subsetneq J^k$. Since the hypercube is defined on the dimensions of $J^k$, we have $x \in A_{J^k}(f)$ and know that it is no use selecting dimensions outside of $J^k$. We conclude that $J^k$ is the unique minimal subset verifying the functionality property, and in general that \textbf{any set of centroids forming an hypercube defined on the dimensions indexes $J$ and with labels corresponding to the coloring of the graph will have the unique selection solution $J$ for all its vertices}.

\subsection{Hypercubes superposition}

We have found a way to create global unique selection solution using hypercubes. We can then superpose several different hypercubes in $\mathcal{X}$. We avoid interactions between hypercubes by storing the coordinates occupied by each hypercube $k$ on each dimensions (\textit{e.g.} in figure \ref{fig:2_14_a}, the bidimensional hypercubes \textit{occupies} the coordinates $\{ -0.375, -0.125 \}$ on the axis $x_1$ and $\{ 0, 0.25 \}$ on axis $x_2$), and ensuring that others allocate different coordinates (\textit{e.g.} in figure \ref{fig:2_14_a}, the univariate hypercube \textit{occupies} the coordinates $\{ 0.125, 0.375 \}$ on $x_1$ and $\{ -0.25 \}$ on $x_2$, which does not collide with the other hypercube).

\subsection{Centroids minimum relative distance}

\begin{figure}[h]
  \centering
  \includegraphics[trim={0 0 1cm 1cm},clip,width=\linewidth]{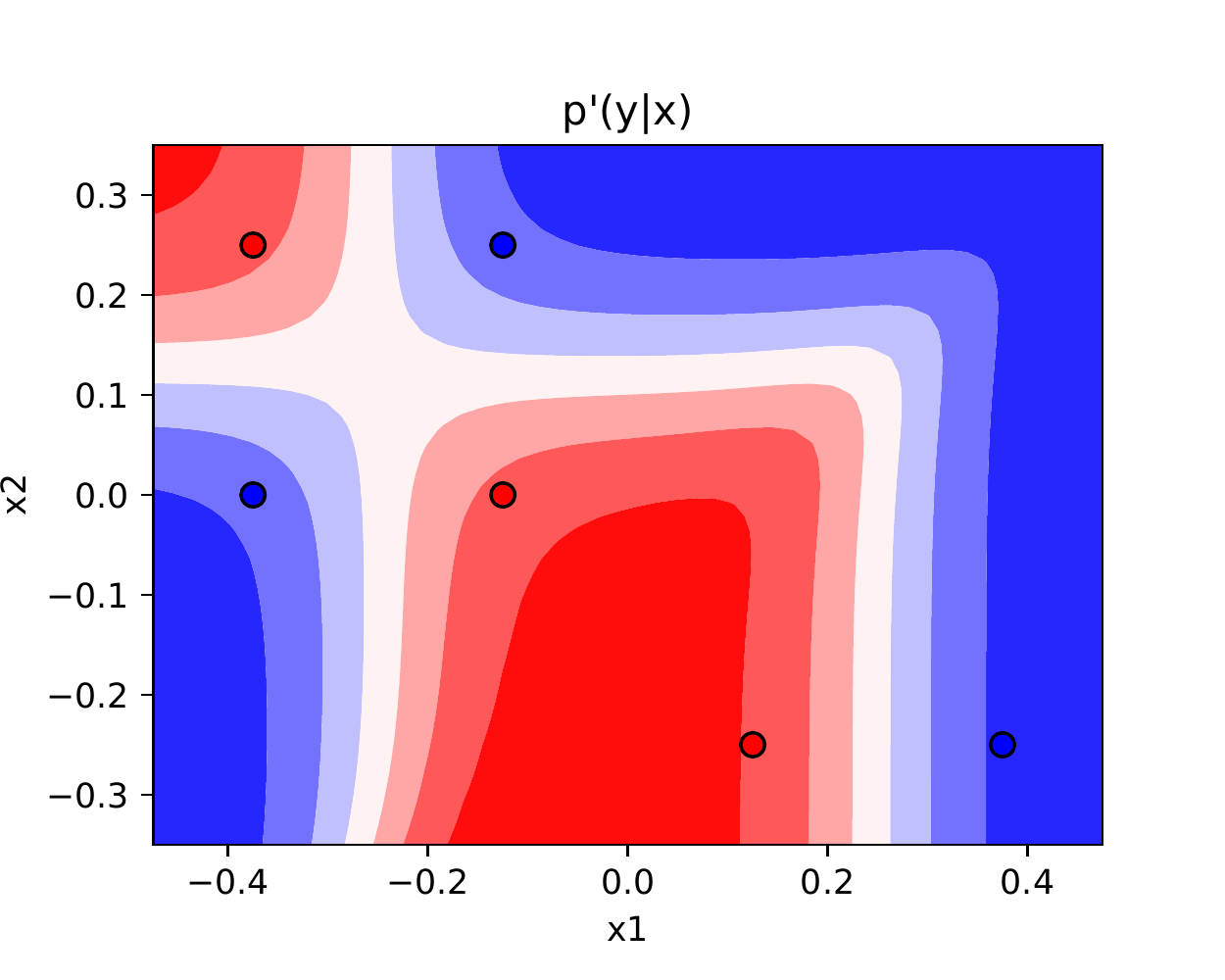}
  \caption{\textbf{Optimal mapping for task \texttt{2\_14}} We plot the conditional probability corresponding to figure \ref{fig:2_14_a} with variance $\sigma/2$.} 
  \label{fig:2_14_b}
\end{figure}

To create the distribution with imperfect dependencies, we also ensure that all occupied coordinates are equally spaced with a minimum distance $\sigma$. Thus, when defining the Gaussian mixture $p'_{X,Y}$ using the centroids as means, we are able to choose the global standard-deviation as a multiple of $\sigma$ (typically $\sigma/2$) to control the superposition ratio of the Gaussian distributions.
With our previous example, the obtained optimal mapping $f$ is shown in figure \ref{fig:2_14_b}.

\subsection{Hypercube erosion}

Lastly, we wanted to create more diversity and break the global ground-truth selection within each hypercube. For that, we randomly erase some centroids in each hypercube with a fixed probability $P_e$. For each centroids $c_j$ we denote the set of its remaining hypercube neighbors $\mathcal{N}_j$. Note that for all $c \in \mathcal{N}_j$,  $c$ differs from $c_j$ by only one coordinate and has an opposite label. Within its hypercube $k$, $c_j$ has its neighbors located on the dimensions $\mathcal{J}_j = \{i \mid \exists c \in \mathcal{N}_j, P_ic_j \neq P_ic \}$. By construction, $\mathcal{J}_j \subset J^k$. Then, by the same reasoning as before, we know that for all $i \in \mathcal{J}_j$, $c_j \in B_i(f)$ and that $c_j \in A_{\mathcal{J}_j}(f)$; and thus deduce that $\mathcal{J}_j$ is the minimal dependence subset for $c_j$ and hence its selection solution $s^*_j$. From this last result, \textbf{a simple principle emerges to visually deduce instance-wise selection for centroids}: we only have to find the set of their hypercube neighbors to deduce the set dimensions indexes containing contradicting labels, then $\mathcal{J}_j = s^*_j$. We have conveniently drawn all neighbor relations in our figures with dotted lines. This last principle is only valid while working with hypercubes.

\begin{figure}[h]
  \centering
  \includegraphics[trim={0 0 0 0.5cm},clip,width=\linewidth]{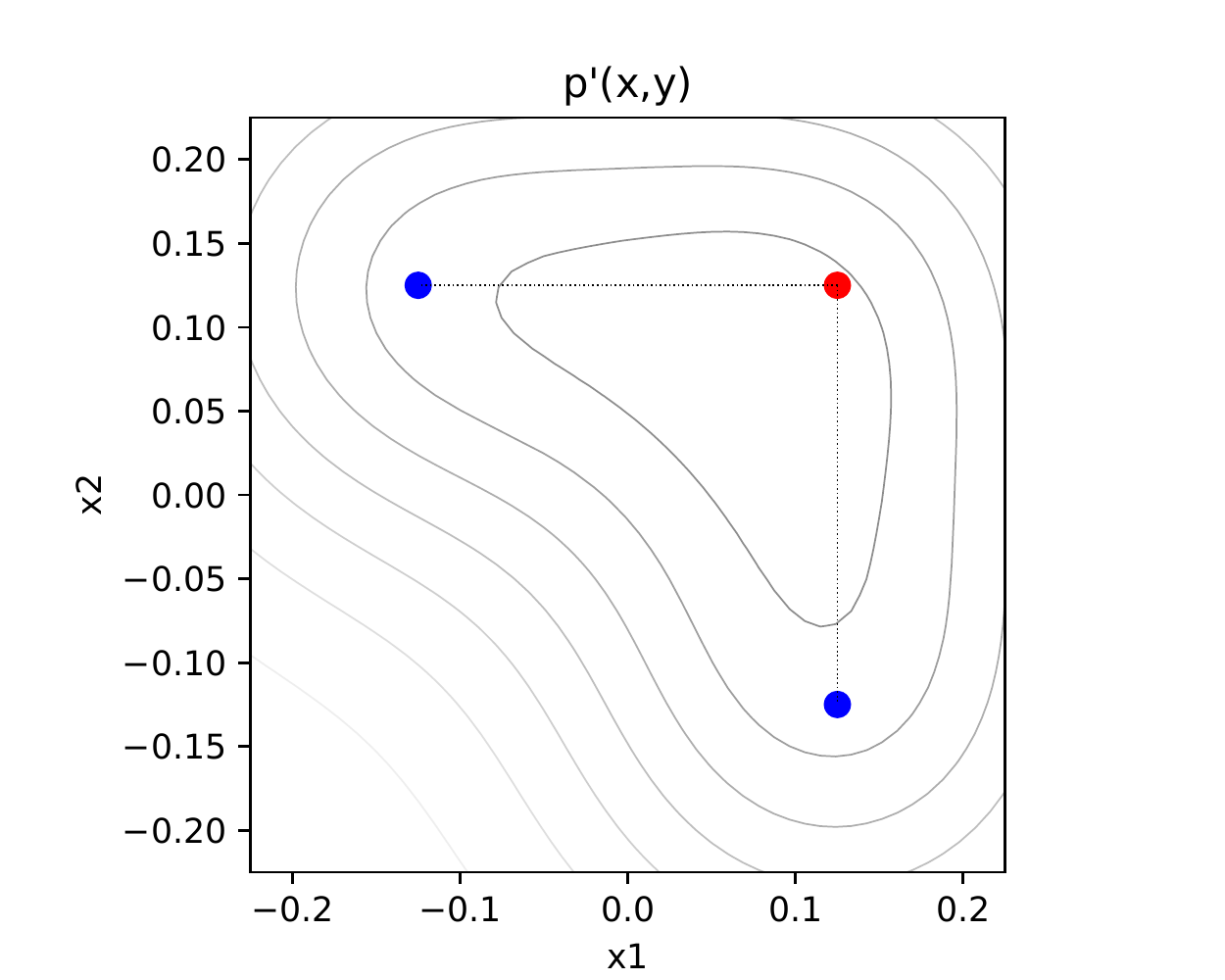}
  \includegraphics[trim={0 0 0 0.5cm},clip,width=\linewidth]{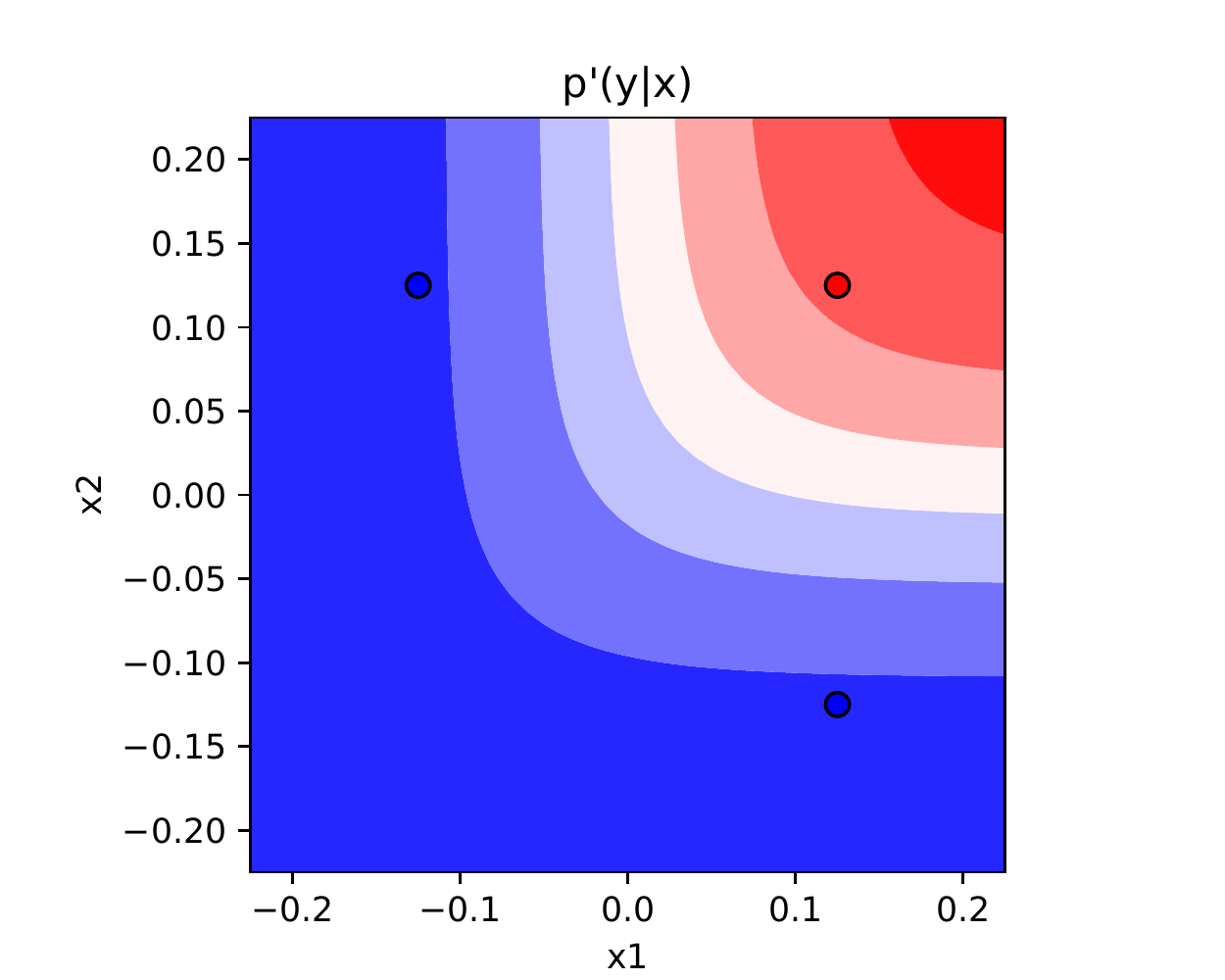}
  \caption{\textbf{Task \texttt{2\_6}} The legend is similar to figure \ref{fig:2_14_a} and \ref{fig:2_14_b}.} 
  \label{fig:2_6}
\end{figure}

We found this "erosion" procedure that deletes random points from an hypercube to be quite interesting as from an initial global selection solution $J^k$ we create many diverse solutions $s^*_j \subset J^k$. An example is given in figure \ref{fig:2_6} where one point was erased from a bidimensional hypercube (\textit{i.e.} a \textit{XOR}). Instead of having a global selection $S^* = \{ 1, 2 \}$, we end up with $s^*_{-0.125, 0.125} = \{ 1 \}$, $s^*_{0.125, 0.125} = \{ 1, 2 \}$ and $s^*_{0.125, -0.125} = \{ 2 \}$.

\subsection{Full generative process}

\begin{figure}[t]
  \centering
  \includegraphics[width=\linewidth]{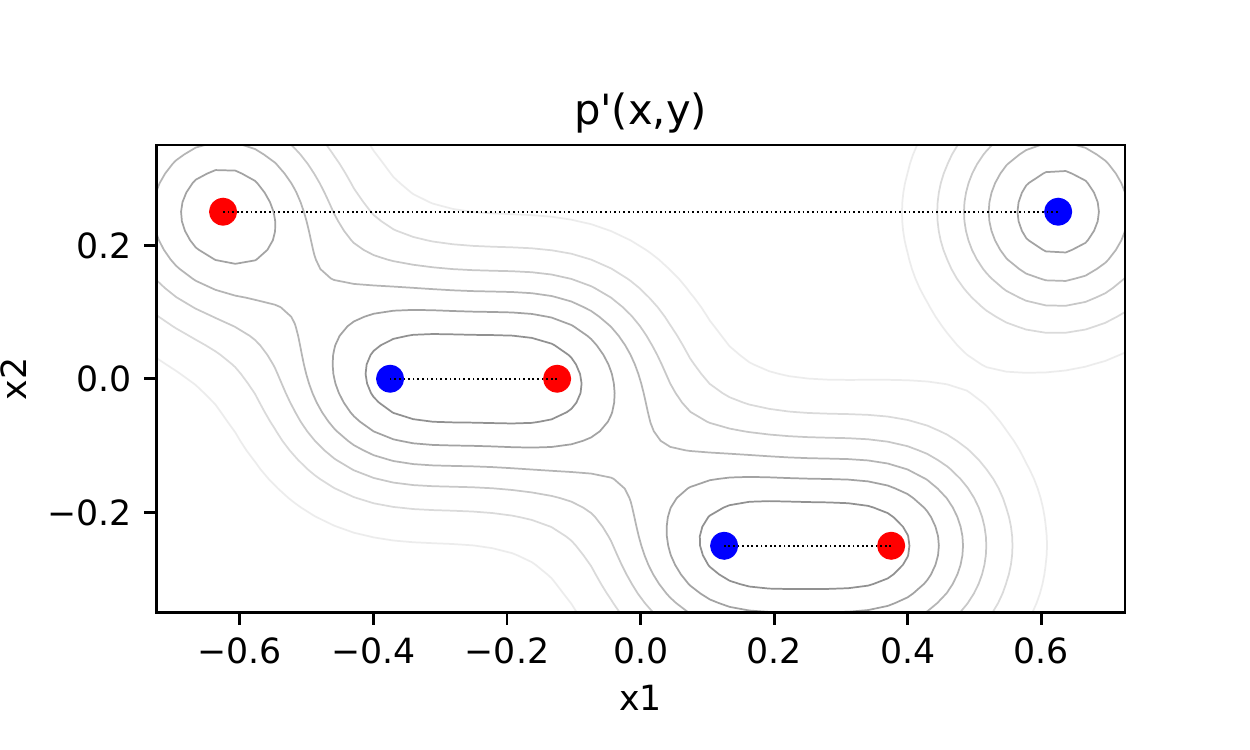}
  \includegraphics[width=\linewidth]{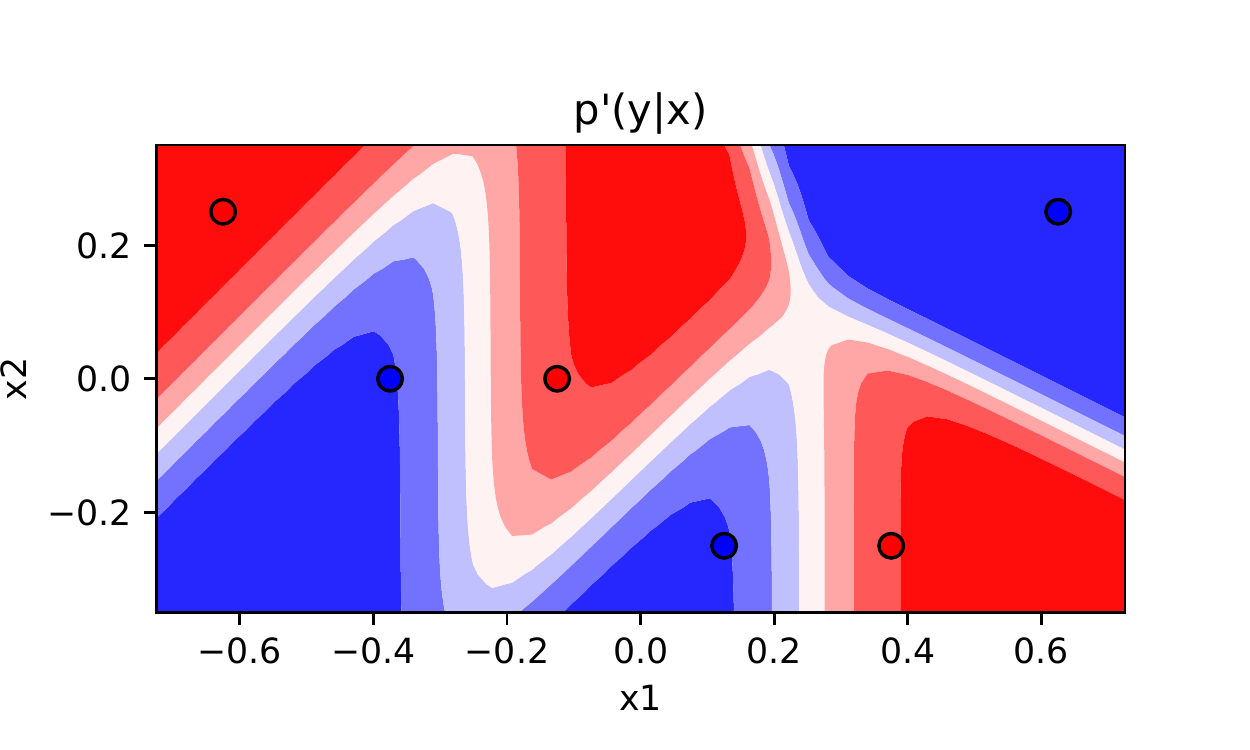}
  \caption{\textbf{Task \texttt{2\_23}} The legend is similar to figures \ref{fig:2_14_a} and \ref{fig:2_14_b}.} 
  \label{fig:2_23}
\end{figure}

\begin{figure}[t]
  \centering
  \includegraphics[trim={0 0 0 1cm},clip,width=\linewidth]{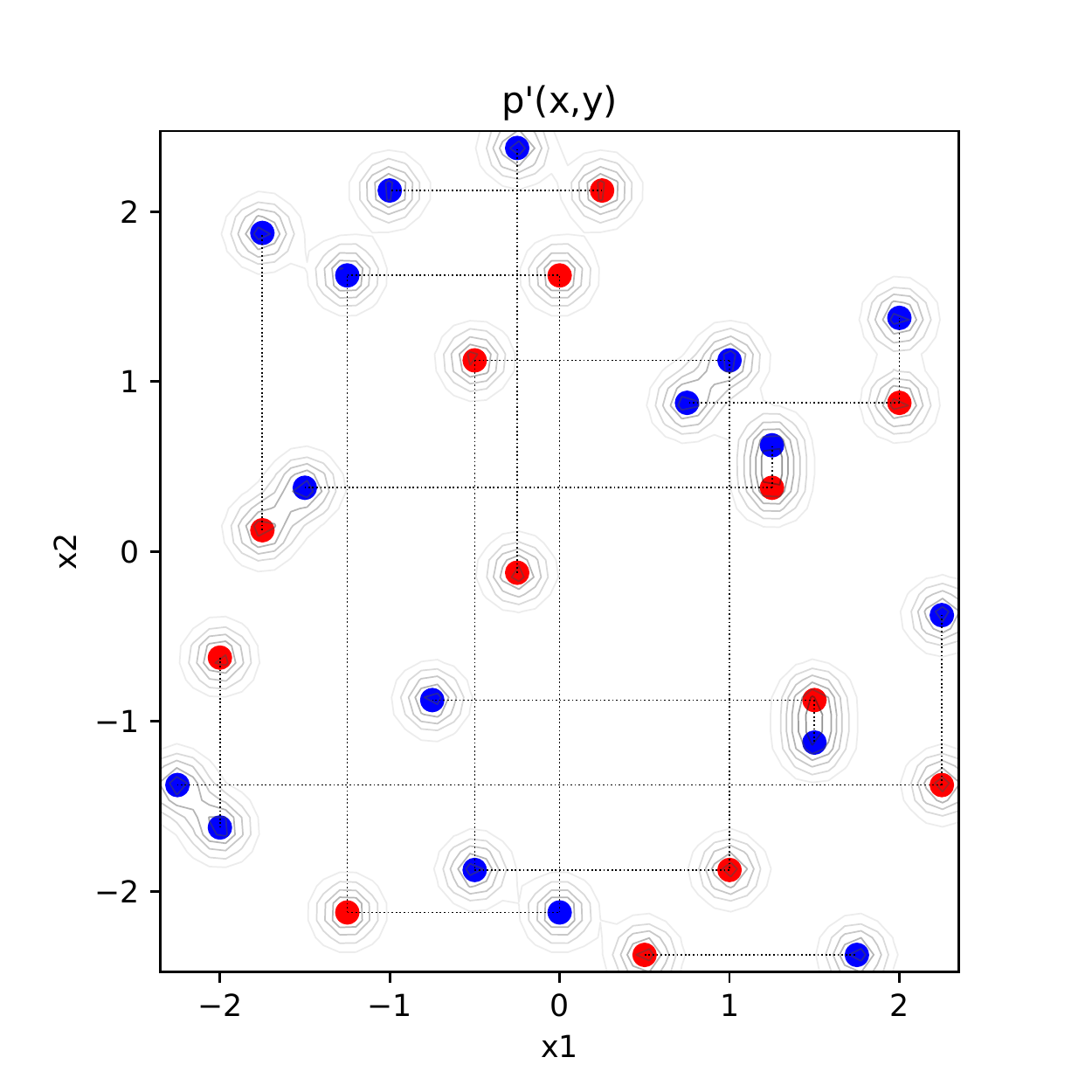}
  \includegraphics[trim={0 0 0 1cm},clip,width=\linewidth]{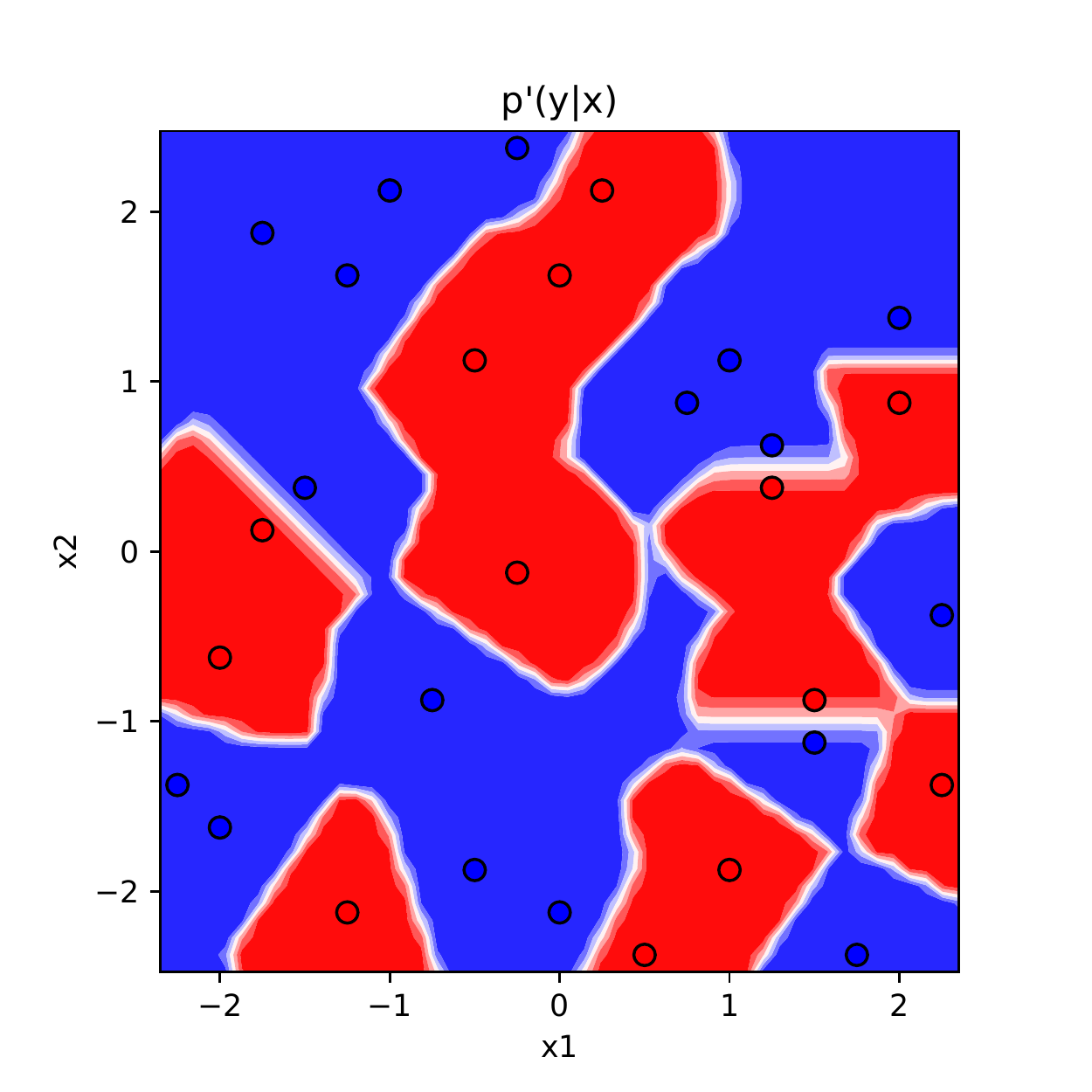}
  \caption{\textbf{Task \texttt{2\_100}} The legend is similar to figures \ref{fig:2_14_a} and \ref{fig:2_14_b}.} 
  \label{fig:2_100}
\end{figure}

\begin{figure}[t]
  \centering
  \includegraphics[trim={2cm 1cm 0 2cm},clip,width=0.9\linewidth]{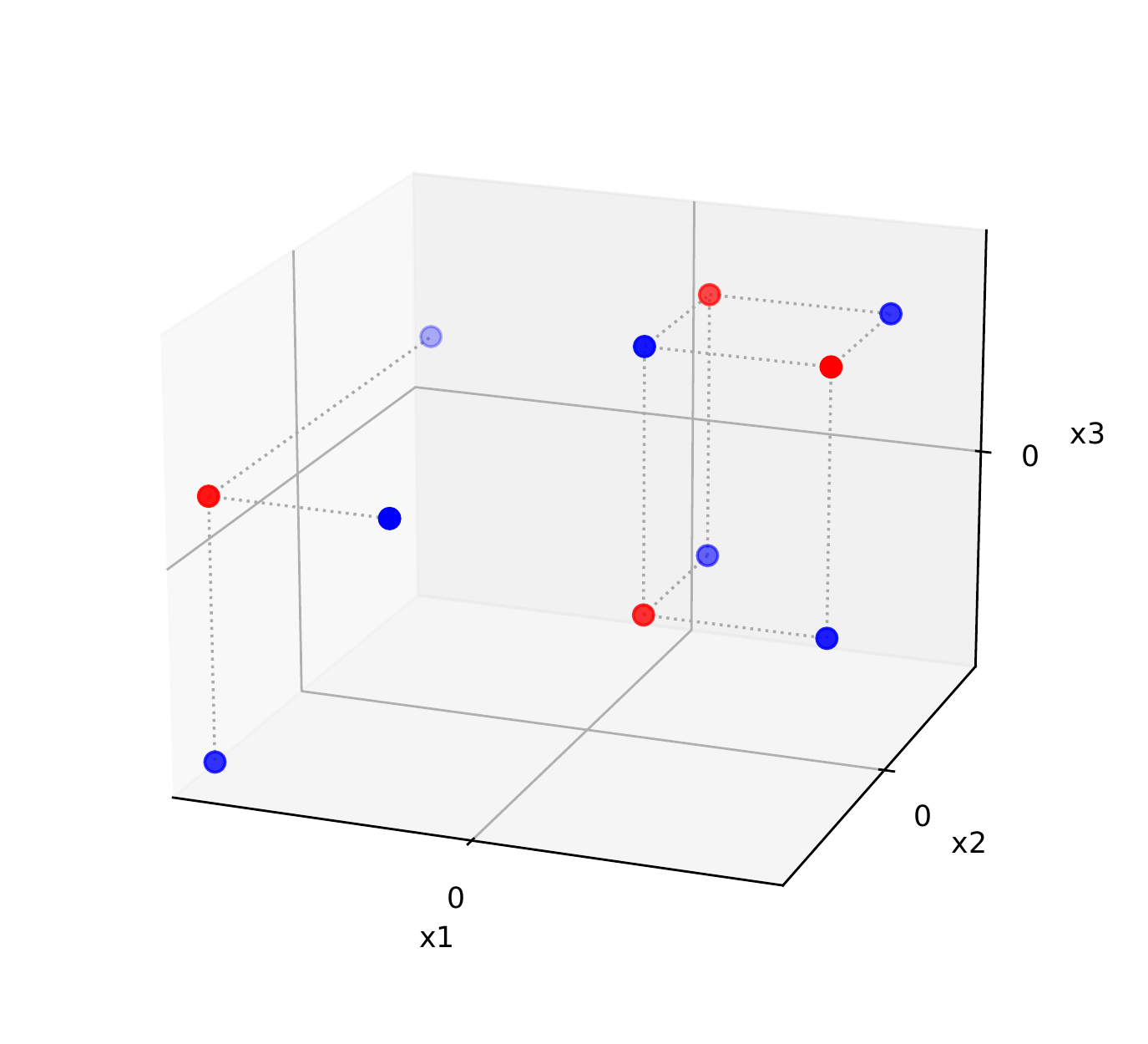}
  \caption{\textbf{Task \texttt{3\_12}} We only display centroids and neighbors} 
  \label{fig:3_12}
\end{figure}

\begin{figure}[t]
  \centering
  \includegraphics[trim={2cm 1cm 0 2cm},clip,width=0.9\linewidth]{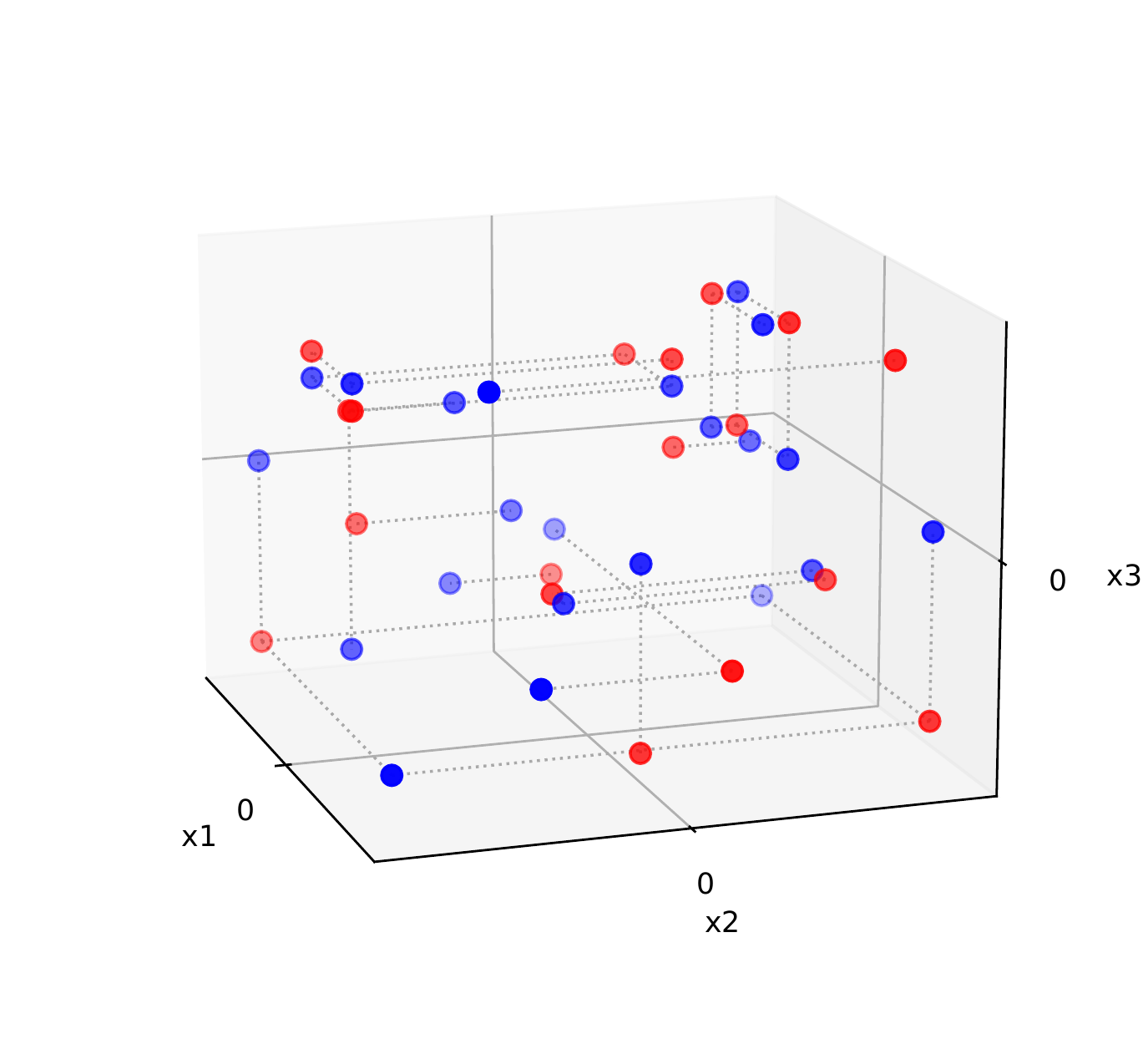}
  \caption{\textbf{Task \texttt{3\_100}}} 
  \label{fig:3_100}
\end{figure}

To create a collection of tasks, we sample a number of hypercubes to create, generate each one by sampling $J^k$ that gives its orientation along the dimensions, and its occupied coordinates, and finally, we randomly erase some points of the hypercube and update $S^*$ accordingly. The corresponding algorithm is provided in Python in the code repository. More generated examples are given in figures \ref{fig:2_23} and \ref{fig:2_100}, as well as examples for $\mathcal{X} \subset \mathbb{R}^3$ in figure \ref{fig:3_12} and \ref{fig:3_100}.

\begin{figure}[t]
  \centering
  \includegraphics[trim={0 1cm 1cm 2cm},clip,width=0.9\linewidth]{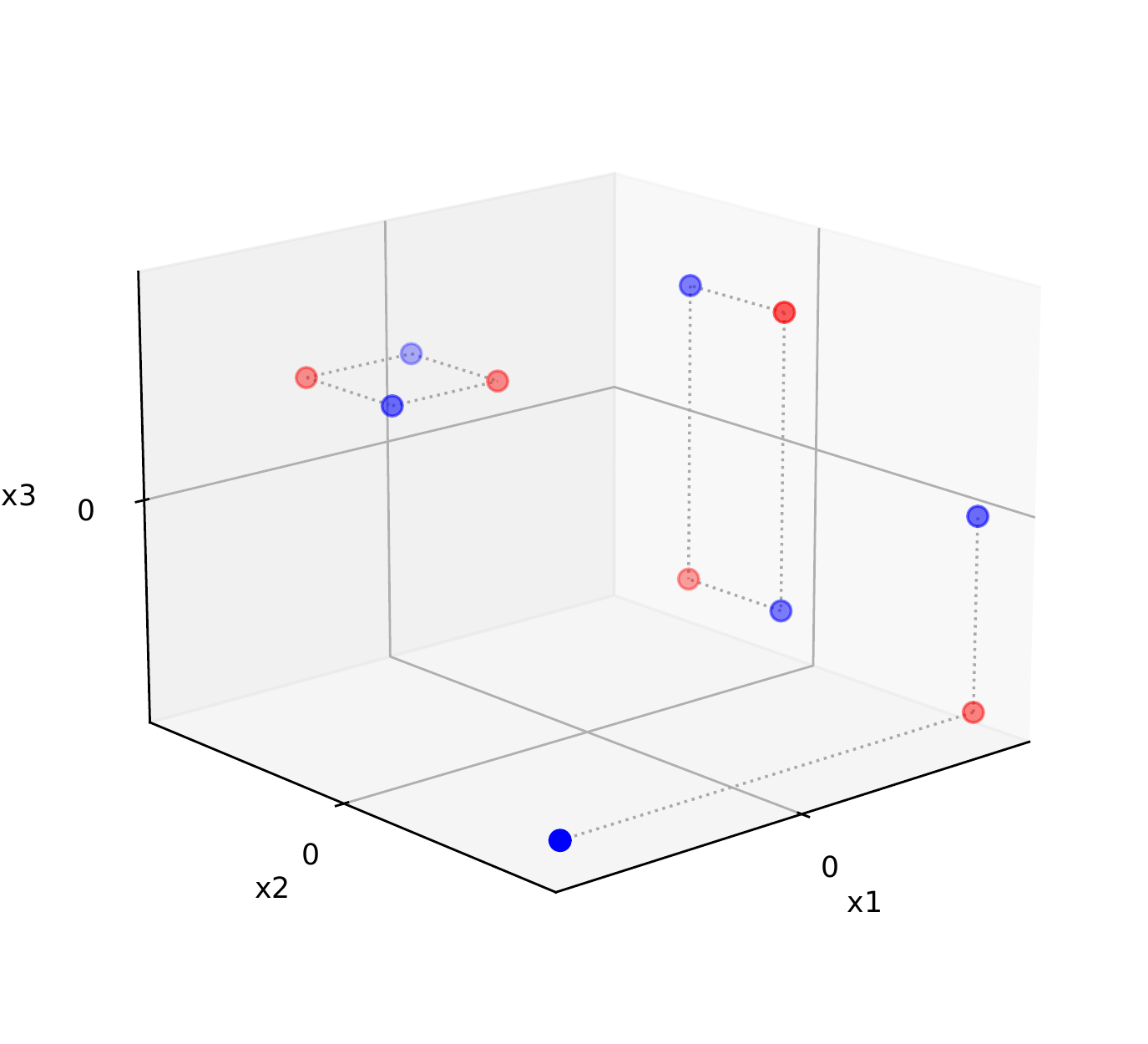}
  \caption{\textbf{Task \texttt{3\_27}}} 
  \label{fig:3_27}
\end{figure}

\section{Experimental details}
\label{annex:exp}

We list more implementation details, configurations and tuned parameters for the methods we evaluate.

\textbf{LIME \cite{ribeiro2016should}}

We set the sampling number to \textbf{1000} and the ridge regression parameter to \textbf{1}, as in the official implementation. Everything else is similar to the official repository.

\textbf{Shapley Sampling}

We set the maximum sampling number to \textbf{128}, meaning that up to dimension 7, we compute exact shapley values, and sample permutations otherwise. We tried augmenting this number to \textbf{512}, which did yield a non-significative \textbf{0.2} accuracy improvement that we chose not to report due to the important trade-off in computation time. We wanted to keep the table clear with a computation budget comparable to \textit{GA\textsuperscript{$\infty$}M} that achieved similar performances.

\textbf{GAM, ... GA\textsuperscript{$\infty$}M}

No hyper-parameter. These methods directly use the conditional probabilities $p(y=1 \mid x_I)$, for which we have access to an analytical form, to estimate each restricted expert $f_I$. Then we use the categorical attribution measure (3), which translates in our case as $\textrm{attr}_I(x) = \max(f_I(x), 1 - f_I(x))$.

\textbf{Grad, Grad $\times$ Input \cite{simonyan2013deep}}

No hyper-parameter. We use noise-free analytical expressions for $f$ and $\nabla f$.

\textbf{Integrated Gradient \cite{sundararajan2017axiomatic}}

We set the sampling number to \textbf{50} for the integral estimation. The baseline point is chosen as the mean of the task centroids.

\textbf{Expected Gradient \cite{erion2019learning}}

We set the sampling number to \textbf{500} for tuples of $\alpha$ interpolation coefficients and background points taken among the task centroids.

\textbf{SHAP \cite{NIPS2017_7062}}

We implement SHAP similarly to Shapley Sampling, the only difference is in the choice of the baseline value. With the original paper notation, $f_S(x) = f(x_S, \mathbb{E}_{\bar{S}}[x_{\bar{S}}])$. Then we directly use $\textrm{attr}_I(x) = \phi_I(x)$.

\textbf{Archipelago \cite{tsang2020does}}

No hyper-parameter.

\textbf{InterpretableNN \cite{afchar2020making}}

With the original paper notation, we choose $g^i_\theta(x) = 4(F^i_\theta(x) - 0.5)^2$, with $F$ our model output in $[0, 1]$. 

\textbf{L2X \cite{chen2018learning}}

We instantiate two three-layer neural network identical in architecture with \texttt{selu} activations, and \textbf{100} neurons in their hidden layers. For the concrete sampler, we chose \textbf{$\tau = 0.1$}, as in the official implementation. We train the predictor with a cross-entropy loss. The whole model is trained for a maximum of \textbf{500} epochs of \textbf{100} steps with a batch-size \textbf{512}. We add an early stopping after \textbf{200} epochs with patience \textbf{10} on the selection solution for the task centroids.

\textbf{INVASE \cite{yoon2018invase}}

We instantiate two three-layer neural network with \texttt{selu} activations, and \textbf{100} neurons in their hidden layers. Following the official implementation, we add batch-normalisation layers in the predictor model.
For the selector -- referred to as "actor" in the original paper, we grid-searched the regularisation parameter in $[0.005, 0.01, 0.02, 0.05, 0.1, 0.2, 0.5, 1]$ and found the value $\lambda^* = 0.01$. The whole model is trained with the same epoch configuration as \textit{L2X}.

\begin{figure}[h]
  \centering
  \includegraphics[width=\linewidth]{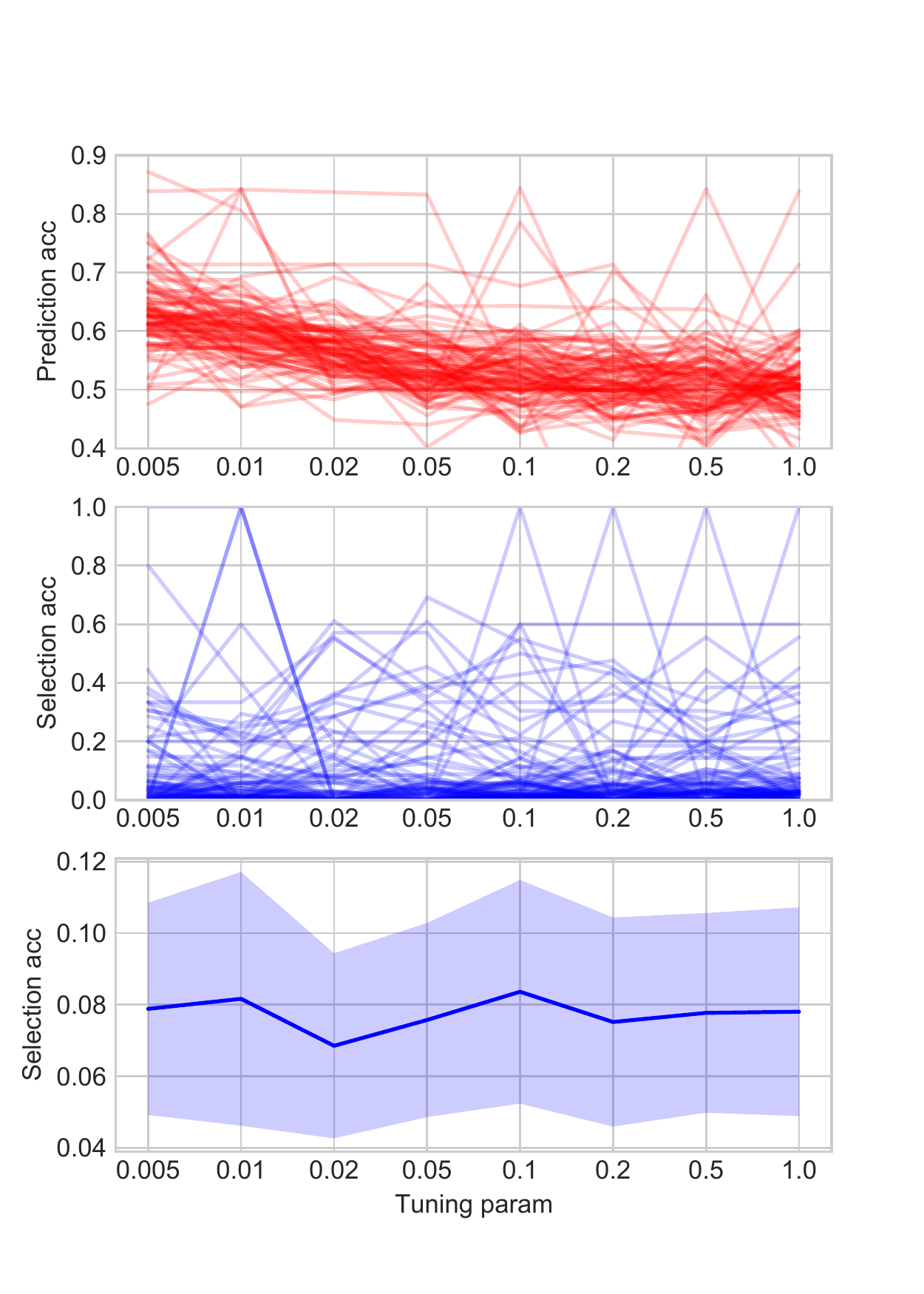}
  \caption{\textbf{Tuning curves for \textit{INVASE}} We plot the \textit{predictor} performances at predicting centroid labels in red, and the \textit{selector} estimated selection mask accuracies in blue. We superpose the tuning curves from each task in the tuning task set and display a clearer aggregated selection accuracy curve with 95\% confidence intervals. Our chosen $\lambda^*$ maximises upper confidence bound.} 
  \label{fig:invase_tuning}
\end{figure}

Tuning $\lambda$ is quite difficult as performances between tasks tend to vary widely with \textit{INVASE}. As illustrated in figure \ref{fig:invase_tuning}, there is no clear significantly better parameter choice. With small $\lambda$ values, the predictor performances increase as the selector tend to select almost all input features; with higher values the selector sparsity constraint dominates and the predictor quickly collapses to a random 50\% accuracy. Most of the time though, \textit{INVASE} does not return the correct selection solution, no matter the value of $\lambda$, we suspect that the method falls into the label-passing trap we covered in section \ref{annex:selpred}.  As we had decided to only grid-search one $\lambda$ value for all tasks -- which worked reasonably well with other methods -- we did not try to further tune this method.

\textbf{Attribution threshold}

\begin{figure}[h]
  \centering
  \includegraphics[width=\linewidth]{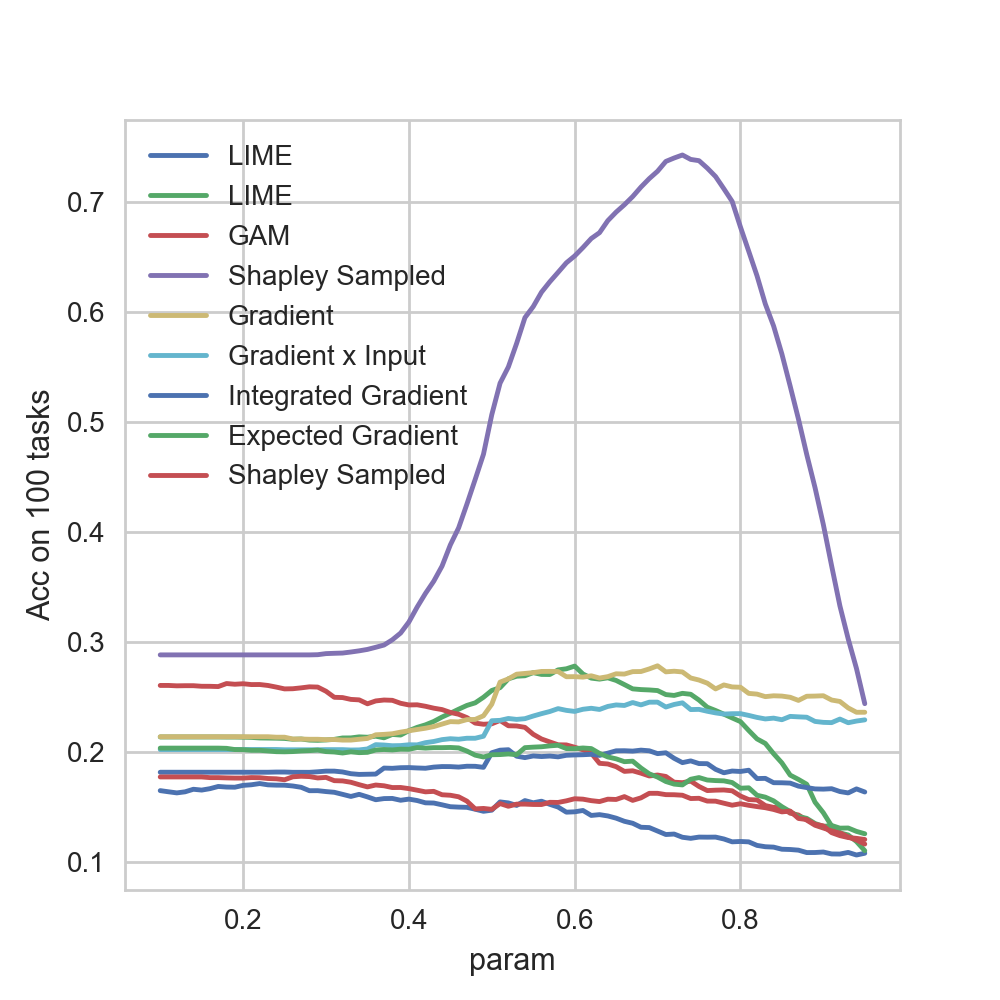}
  \includegraphics[width=\linewidth]{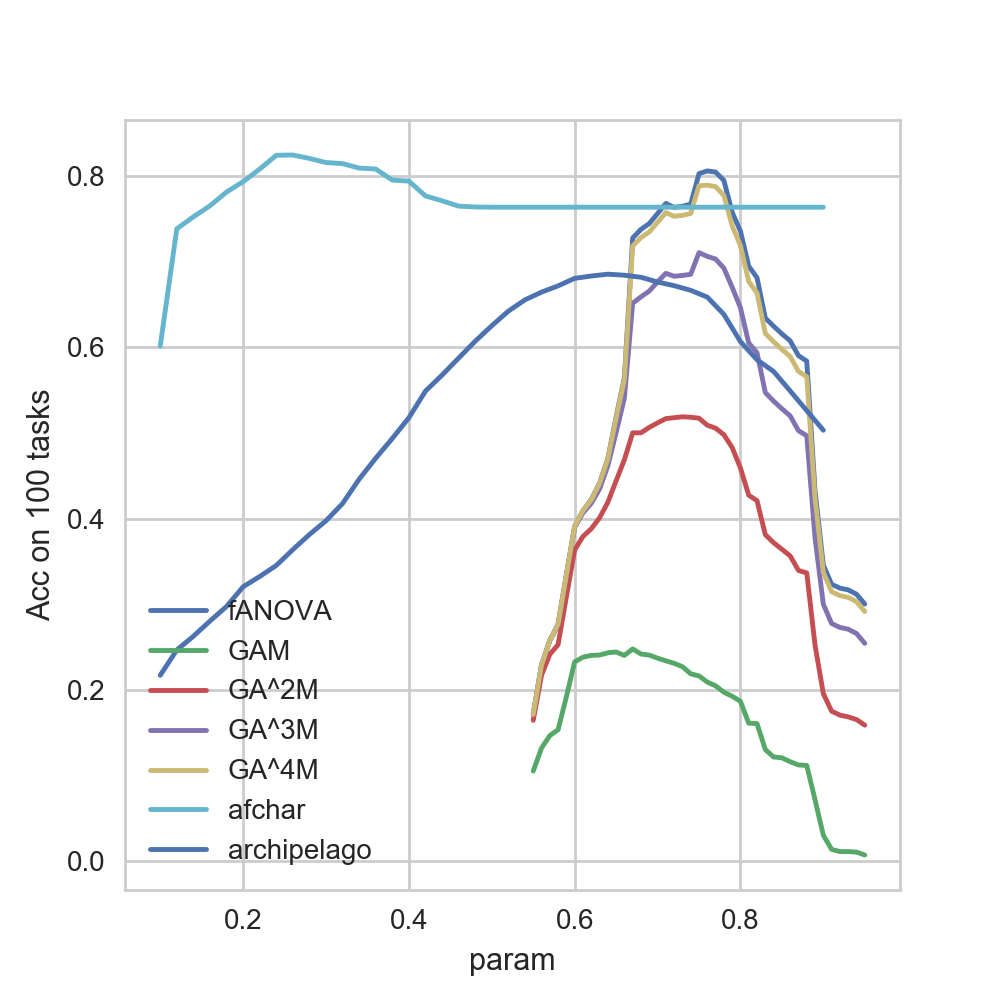}
  \caption{Tuning curves of subset attribution methods} 
  \label{fig:tuning_curves}
\end{figure}

For all methods returning \textbf{feature} attributions \textit{i.e.} $n$ values for the $n$ input features, given values for each point, we select all features with an attribution value higher than $\mu$ times the maximum attribution on this point. We tune this multiplicative coefficient by evaluating the methods on 100 tasks, generated and used only for tuning, on a range $[0.1, 0.95]$ with step $0.01$. Except for selector-predictors, we have observed rather convex curves of performances and clear maximums for each method, as displayed in figure \ref{fig:tuning_curves}. The obtained parameters are given in table \ref{tab:tuned_features}. We must underline that the variety in the found coefficients support the specificity and task-dependence we mentioned for each method in the definition of their attribution relative values: some yield sparse attributions values, others do not.

\begin{table}[h]
\centering
\begin{tabular}{c|c}
    \toprule
    Method & $\mu^*$ \\
    \midrule
    LIME (Cat.) & 0.23 \\
    LIME (Cont.) & 0.61 \\
    GAM & 0.18 \\
    Shapley ($\mathbb{E}(f)$) & 0.73 \\
    SHAP ($f(\mathbb{E})$) & 0.28 \\
    Gradient & 0.67 \\
    Gradient x Input & 0.73 \\
    Integrated Gradient & 0.52\\
    Expected Gradient & 0.56 \\
    \midrule
    L2X & 0.82 \\
    INVASE & 0.55 \\
    \bottomrule
\end{tabular}
\caption{Tuned multiplicative coefficient $\mu$ to estimate subset selection from feature attribution.}
\label{tab:tuned_features}
\end{table}

The remaining methods return \textbf{subset} attribution values -- \textit{i.e.} $2^n$ values.
They provide estimations of many conditional means $\mathbb{E}(Y \mid X_I = x_I)$, up to a fixed cardinality for $I$ -- two for GA\textsuperscript{2}M, three for GA\textsuperscript{3}M, etc.
In our case, this last quantity is directly equal to $\mathbb{P}(Y = 1 \mid X_I = x_I)$, and we thus obtain the simple attribution measure (3) we derived by applying the function $g(p) = \max(p, 1-p)$ on each subfunction output.
Then, we use a threshold parameter $\eta$ and find the subset $I$ with lowest cardinality such that $\textrm{attr}_I(x) > \eta$. As we have a binary problem, $\eta$ is bounded in $[1/2, 1]$, we tune $\eta$ in this range with $0.01$ steps. For \textit{InterpretableNN}, a custom function is applied over the probability and yield a method-specific attribution value in $[0, 1]$, we tune $\eta$ in this range with steps $0.01$. The results are given in table \ref{tab:tuned_subset}.

\begin{table}[h]
\centering
\begin{tabular}{c|c}
    \toprule
    Method & $\eta^*$ \\
    \midrule
    fANOVA & 0.76 \\
    $GA^2M$ & 0.75 \\
    $GA^3M$ & 0.75 \\
    $GA^4M$ & 0.76 \\
    \midrule
    Archipelago & 0.66 \\
    InterpretableNN & 0.26 \\
    \bottomrule
\end{tabular}
\caption{Tuned multiplicative coefficient $\mu$ to estimate subset selection from feature attribution.}
\label{tab:tuned_subset}
\end{table}

\textbf{Training}

Most models use analytical expressions of $p'_{X,Y}$ and can be evaluated on a consumer grade computer on CPU in less than an hour for each task set. Selector-predictors require a full training procedure and were trained in parallel on four \textit{GeForce GTX 1080} GPUs. Reported running times are aggregated.

\section{Issues with model-based interpretations}
\label{annex:trees}

Following the reviewing process of our paper, we have decided to add a discussion on the comparison/applicability of our formalisation to feature-based interpretations methods that rely on the inspection of the internal of trained models.

As first remark, though we only work on synthetic data distribution $p'(y|x)$ in our experiment section, our framework is perfectly applicable to a model induced distribution $p_\theta(y|x)$. As mentioned in introduction, this can make evaluation trickier with the added difficulty of disentangling model prediction errors from interpretation errors on unlabeled data, and requires to manage out-of-distribution artifacts impacting the produced interpretations, which is also the case of models architecture and hyperparameters choice \cite{dombrowski2019explanations, kumar2020problems, slack2020fooling}.
The approach in itself would however remain \textit{model-agnostic} and solely inspect the learnt input-output association. This is arguably a common principle in the interpretation field (\textit{e.g.} \cite{ribeiro2016should, NIPS2017_7062}), but a concern was raised of whether inspecting trained weights would not simplify the attribution problem.

For instance, in the case of a decision tree -- that are often considered one of the models with the highest transparency level \cite{arrieta2020explainable} -- a commonly used principle to find responsible features is to aggregate encountered features on which the branching are done from root to leaf to form a prediction rationale. The interpretability of trees and other simple models has already been disputed before \cite{lipton2018mythos, dinu2020challenging}; here, to fix ideas, we highlight an example where the two explanation principles would differ.
Consider figure \ref{fig:2_6}: on one side, we have seen that our method allows to derive a \textit{unique} minimal instance-wise selection solution. Conversely, though the tree-based suggestion seems reasonable at first glance, there exists two optimal trees classifying all three clusters perfectly,
\begin{align*}
    T_1(x) := & \texttt{ if } x_1 < 0 \texttt{ then } 0 \\
    & \texttt{ else } (\texttt{if } x_2 > 0 \texttt{ then } 1 \texttt{ else } 0) \\
    T_2(x) := & \texttt{ if } x_2 < 0 \texttt{ then } 0 \\
    & \texttt{ else } (\texttt{if } x_1 > 0 \texttt{ then } 1 \texttt{ else } 0)
\end{align*}
leading to \textit{two contradicted selections that may be equiprobably returned} on several runs:
\begin{table}[H]
    \centering
    \begin{tabular}{c|cc|c}
        $(x_1, x_2)$ & $T_1$ & $T_2$ & \textbf{Our}  \\
        \toprule
        -0.125, 0.125 & $\{ 1 \}$ & $\{ 1, 2 \}$ & $\{ 1 \}$ \\
        0.125, 0.125 & $\{ 1, 2 \}$ & $\{ 1, 2 \}$ & $\{ 1, 2 \}$ \\
        0.125, -0.125 & $\{ 1, 2 \}$ & $\{ 2 \}$ & $\{ 2 \}$ \\
    \end{tabular}
\end{table}
Trees suffer from \textit{identifiability} issues leading to unstable explanations, which seems unsuitable to gain general knowledge. The clusters symmetry between $X_1$ and $X_2$ also seems a good argument in favor of our solution.

Leveraging model inner mechanisms for interpretation has lead to many well performing algorithms, but this also paves the ways to many undesired side-effects that are not immediately visible when working with a high-performing trained model. Our message is that prediction performance and computation transparency does not necessarily translate into interpretation performance. Optimistically, we believe that the study we have conducted on synthetic data helps find those inconsistencies and failure points and may lead to better behaved interpretable models.

\end{document}